%% file: main.tex
\documentclass[10pt,twocolumn,letterpaper]{article}

\usepackage{iccv}
\usepackage{times}
\usepackage{epsfig}
\usepackage{graphicx}
\usepackage{amsmath,bm}
\usepackage{amssymb}

\usepackage[pagebackref=true,breaklinks=true,letterpaper=true,colorlinks,bookmarks=false]{hyperref}

\iccvfinalcopy 

\def\iccvPaperID{10333} 

\ificcvfinal\fi

\begin{document}

\title{SC-NeuS: Consistent Neural Surface Reconstruction \\ from Sparse and Noisy Views }

\author{Shi-Sheng Huang\\
Beijing Normal University \\
{\tt\small huangss@bnu.edu.cn}
\and
Zi-Xin Zou\\
Tsinghua University\\
{\tt\small zouzx19@mails.tsinghua.edu.cn} 
\and
Yi-Chi Zhang\\
Beijing Institute of Technology\\
{\tt\small zhangyc@bit.edu.cn} 
\and
Hua Huang\thanks{corresponding author.}\\
Beijing Normal University \\
{\tt\small huahuang@bnu.edu.cn}
}

\maketitle

\begin{abstract}
   The recent {neu}ral {s}urface reconstruction approaches using volume rendering have made much progress by achieving impressive surface reconstruction quality, but are still limited to dense and highly accurate posed views. To overcome such drawbacks, this paper pays special attention on the consistent surface reconstruction from sparse views with noisy camera poses. Unlike previous approaches, the key difference of this paper is to exploit the multi-view constraints directly from the explicit geometry of the neural surface, which can be used as effective regularization to jointly learn the neural surface and refine the camera poses. To build effective multi-view constraints, we introduce a fast differentiable on-surface intersection to generate on-surface points, and propose view-consistent losses on such differentiable points to regularize the neural surface learning. Based on this point, we propose a joint learning strategy for both neural surface representation and camera poses, named SC-NeuS, to perform geometry-consistent surface reconstruction in an end-to-end manner. With extensive evaluation on public datasets, our SC-NeuS can achieve consistently better surface reconstruction results with fine-grained details than previous state-of-the-art neural surface reconstruction approaches, especially from sparse and noisy camera views. The source code is avaiable at \url{https://github.com/zouzx/sc-neus.git}. 
\end{abstract}

\begin{figure}[t]
\begin{center}
   \includegraphics[width=0.95\linewidth]{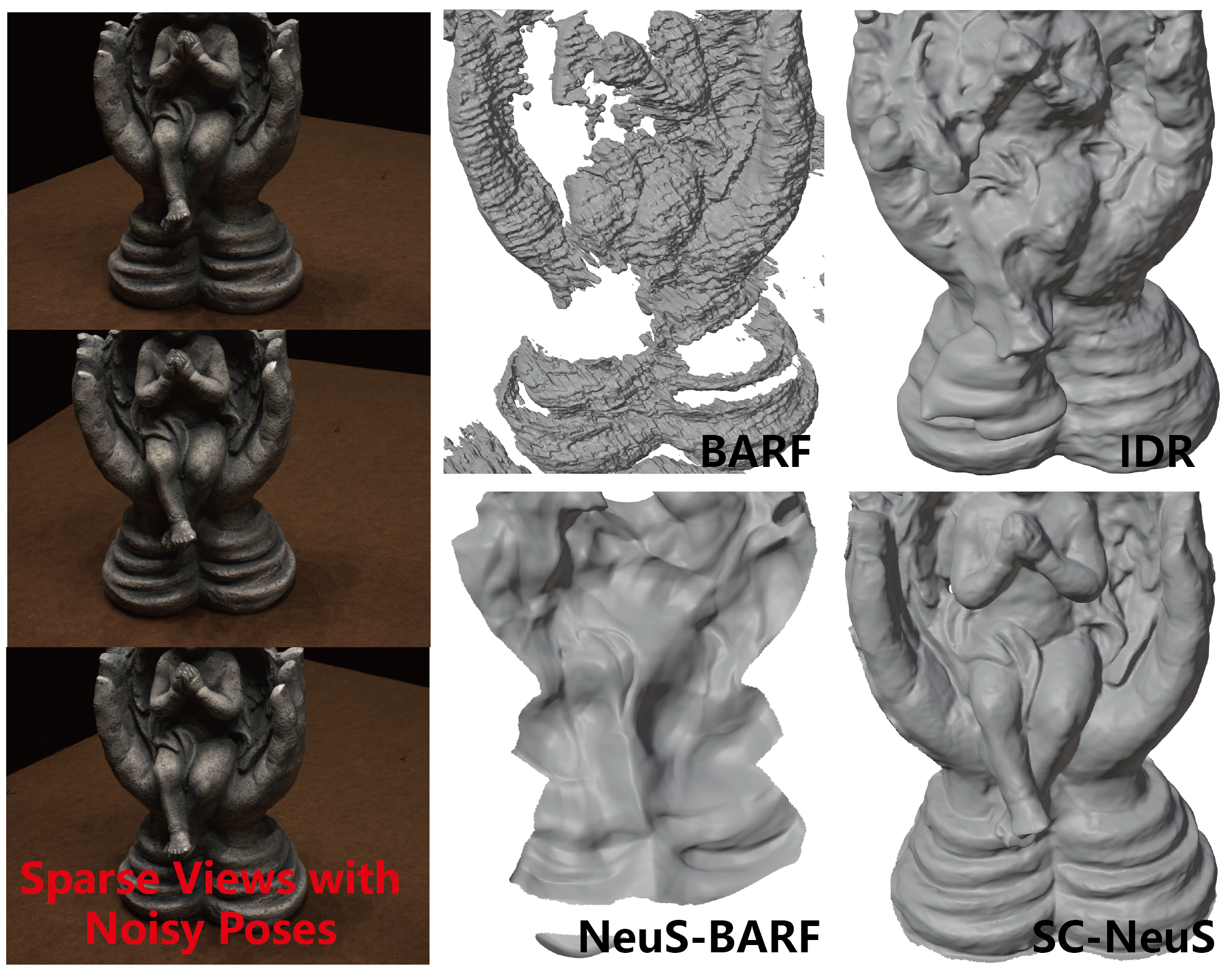}
\end{center}
   \caption{\textbf{Surface reconstruction results from sparse input images with noisy camera poses}. Although enabling neural surface learning from dense input views with noisy camera poses, the current state-of-the-art approaches like BARF~\cite{lin2021barf}, IDR~\cite{yariv2020multiview} still struggle to achieve geometry-consistent surface reconstruction from sparse input views. Instead, our SC-NeuS explores the multi-view constraints directly from the explicit geometry of neural surface, and achieves geometry-consistent surface reconstruction results with fine-grained geometric details. Note that NeuS~\cite{wang2021neus} with the coarse-to-fine learning of BARF~\cite{lin2021barf} (NeuS-BARF) also fails to achieve satisfactory results.  }
\label{teaser}
\end{figure}

\section{Introduction}\label{sec_intro}
3D surface reconstruction from multi-view images continues to be an important research topic in computer vision and graphics communities. Unlike traditional Multi-View Stereo (MVS) based methods leveraging structure from motion (S\emph{f}M)~\cite{snavely2006photo} technique for sparse~\cite{labatut2007efficient,schonberger2016pixelwise,schonberger2016structure,xu2019multi} or dense~\cite{kar2017learning,yao2018mvsnet,xu2020pvsnet} surface reconstruction, the recent neural surface reconstruction approaches~\cite{yariv2020multiview,wang2021neus,oechsle2021unisurf,azinovic2022neural,darmon2022improving,fugeo} adopt to learn the deep implicit representation~\cite{
park2019deepsdf,peng2020convolutional,atzmon2020sal,jiang2020local} with the aid of volume rendering~\cite{mildenhall2021nerf}, leading to more better \emph{complete} and \emph{fine-grained} surface reconstruction quality,  which have received much research attention for multi-view image based 3D reconstruction.

As like Neural Radiance Fields (NeRF)~\cite{mildenhall2021nerf}, one main drawback of most neural surface reconstruction approaches (NeuS~\cite{wang2021neus}, VolSDF~\cite{yariv2021volume},  Unisurf~\cite{oechsle2021unisurf}, NeuralWarp~\cite{darmon2022improving}, Geo-NeuS~\cite{fugeo}) is the dependency on dense input views, which is not suitable for wide real-world applications with only sparse input views and often noisy camera poses in AR/VR, autonomous driving or robotics. Some subsequent works propose to improve the reconstruction quality from sparse scenarios, by introducing regularization like sparse points~\cite{deng2022depth}, multi-views depth priors~\cite{chen2021mvsnerf,niemeyer2022regnerf}, rendering ray entropy~\cite{kim2022infonerf} or geometry-aware feature volume~\cite{long2022sparseneus}. However, most of these approaches are still relying on highly accurate camera poses, which could not be easily obtained using 
technique like COLMAP~\cite{schonberger2016structure} for sparse input views. 

To overcome the dependency on highly accurate cameras poses, many recent works propose to jointly learn the deep implicit geometry and refine the camera poses, with the guidance of novel registration from photometric~\cite{chng2022gaussian,lin2021barf,meng2021gnerf,wang2021nerfmm,yariv2020multiview} or silhouette~\cite{bosssamurai,kuang2022neroic,zhang2021ners} priors. But since those registrations are often performed \emph{independently} across dense input views, the registration quality would significantly drop for sparse view scenarios (Fig.~\ref{teaser}), where enough \emph{relations} across views are missing to effectively bundle adjust both the deep implicit geometry and camera poses. It still remains to be challenging to jointly learn the deep implicit geometry and camera poses from sparse input views~\cite{zhang2022relpose} for geometry-consistent surface reconstruction.   

This paper proposes a \textbf{S}parse-view \textbf{C}onsisent \textbf{Neu}ral \textbf{S}urface (SC-NeuS) learning strategy, which performs geometry-consistent surface reconstruction with fine-grained details from sparse and noisy camera poses (as few as 3 views). Unlike previous \emph{independent} registrations from dense input views, we seek to explore more effective multi-view constraints between sparse views.
Due to the gap between the volume rendering integral and point-based SDF modeling~\cite{fugeo}, except from relying on the depth constraints~\cite{chen2021mvsnerf} rendered from the under-constrained signed distance field~\cite{fugeo}, we utilize extra regularization directly from the explicit geometry of the neural surface representation.
Our key insight is that the observation of the explicit surface geometry across multiple views should be \emph{consistent}, which can be used as effective regularization to jointly learn both the neural surface representation and camera poses. Specifically, we first introduce a fast \emph{differentiable} on-surface intersection to sample on-surface points from explicit geometry of the neural surface, and then provide effective view-consistent losses defined on such differentiable on-surface intersections, which builds up an end-to-end joint learning for the neural surface representation and camera poses. Besides, to further improve the geometry-consistent neural surface learning, we incorporate an coarse-to-fine learning strategy~\cite{lin2021barf} for highly accurate and fine-grained surface reconstruction results. 

 To evaluate the effectiveness of our SC-NeuS, we conduct extensive experiments on public dataset including DTU~\cite{jensen2014large} and BlendedMVS~\cite{DBLP:conf/cvpr/0008LLZRZFQ20} with various geometry scenarios. Compared with previous state-of-the-art approaches\cite{lin2021barf,fugeo,jeong2021self,yariv2020multiview,wang2021neus}, our SC-NeuS achieves consistently better geometry-consistent surface reconstruction results both quantitatively and qualitatively, which becomes a new state-of-the-art neural surface reconstruction approach from sparse and noisy cameras. 

\begin{figure*}[htbp]
\begin{center}
   \includegraphics[width=1\linewidth]{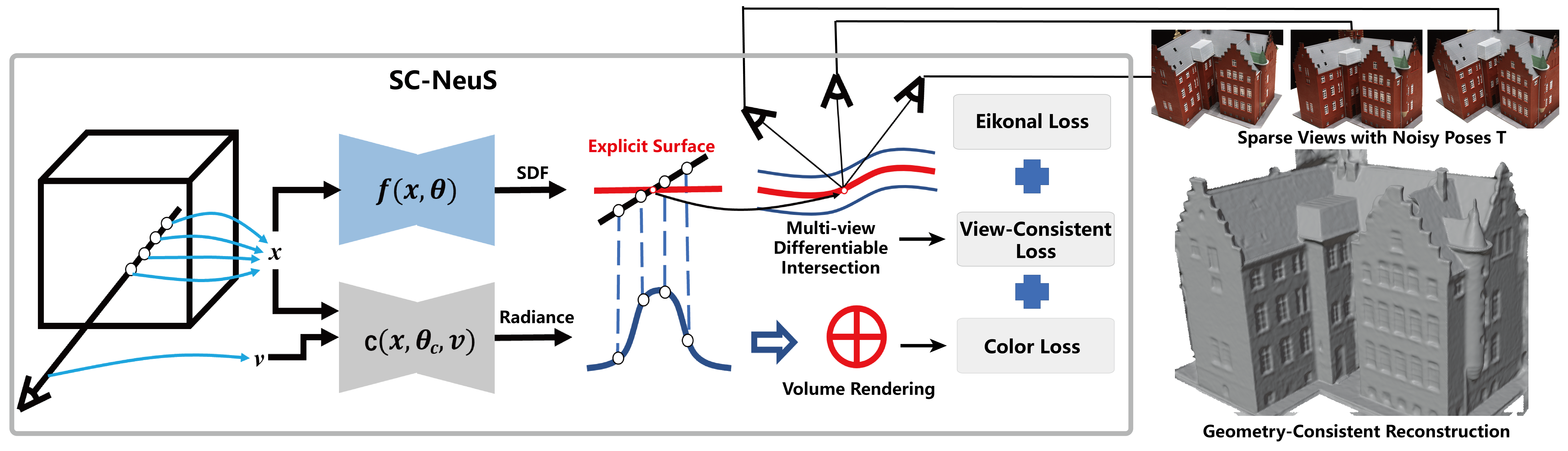}
\end{center}
   \caption{\textbf{The overview of our SC-NeuS}. Given sparse input images (as few as 3 views) with noisy camera poses $T$, our SC-NeuS represents the object's geometry as a signed distance filed $f(x,\bm{\theta})$ and perform volume rendering of the geometry with an extra radiance filed $c(x,\bm{\theta_c},v)$. Combing the effective view-consistent loss defined on the multi-view differentiable intersection points on the explicit surface of $f(x,\bm{\theta})$, with color loss and Eikonal loss, our SC-Neus performs jointly learning of $f(x,\bm{\theta})$, $c(x,\bm{\theta_c},v)$ and camera poses $T$ in an end-to-end manner, achieving geometry-consistent surface reconstruction results with fine-grained details. }
\label{system}
\end{figure*}

\section{Related Work}\label{sec_related_work}

\textbf{Novel View Synthesis.} The recent success of Neural Radiance Fields (NeRF)~\cite{mildenhall2021nerf} has inspired many subsequent works~\cite{trevithick2021grf,wang2021ibrnet,yu2021pixelnerf} to achieve impressive novel view synthesis applications. To overcome the drawback of dense input views, multiple works propose to extra regularization or priors for sparse view novel view synthesis. RegNeRF~\cite{niemeyer2022regnerf} proposes to regularize the rendered patches with depth and appearance smoothness for sparse view synthesis. MVSNeRF~\cite{chen2021mvsnerf} leverages similar rendered depth smoothness loss across unobserved views, from pre-trained sparse view to generalized novel view synthesis. On other hand, InfoNeRF~\cite{kim2022infonerf} penalizeds the NeRF overfitting to limited input views with a ray entropy regularization. Mip-NeRf360~\cite{barron2022mip} introduce ray distortion loss, which encourages sparsity of the density learning in each rendering ray. Besides, some recent approaches~\cite{wei2021nerfingmvs,deng2022depth,roessle2022dense} use depth priors to constraint the NeRF optimization, which also achieves promising novel view synthesis results from sparse input views. Different from all of these previous approaches that relies on highly accurate camera poses as input, our approaches aims at geometry-consistent neural surface learning with noisy camera poses, and contributes a joint neural surface learning and camera pose optimization strategy from sparse input views. 

\textbf{Neural Implicit Surface Representation.} Neural implicit representation has been a state-of-the-art way to represent the geometry of objects or scenes since the pioneer works of DeepSDF~\cite{park2019deepsdf} and its subsequents~\cite{jiang2020local}. IDR~\cite{yariv2020multiview} introduces a neural surface rendering for the neural implicit representation (signed distance function, SDF), which enables precise surface learning from 2D images. Inspired by the success of NeRF~\cite{mildenhall2021nerf}, NeuS~\cite{wang2021neus} and VolSDF~\cite{yariv2021volume} propose to transfer the signed distance field to density filed using weight function, and perform the volume rendering along with the radiance field, achieving impressive surface reconstruction results with fine-grained details. Geo-NeuS~\cite{fugeo} incorperates more explicit surface supervisions for more accurate neural surface learning.  UNISURF~\cite{oechsle2021unisurf} explores the balance between surface rendering and volume rendering. NeuralWarp~\cite{darmon2022improving} provides a geometry-aware volume rendering which utilize multi-view geometry priors for geometry-consistent surface reconstruction. However, most of these previous works depend on dense input views for accurate neural surface learning, which is not feasible for sparse scenarios. 

Recently, SparseNeuS~\cite{long2022sparseneus} learns geometry encoding priors from image features for generalizable neural surface learning form sparse input views, but still relies on highly accurate camera poses. In contrast, our approaches enables accurate neural surface learning from sparse input views, and optimizes the noisy camera poses simultaneously. 

\textbf{Joint Deep Implicit Geometry and Pose Optimization.} BARF~\cite{lin2021barf} is probably one of the first works to reduce NeRF's dependent on highly accurate camera poses, by introducing a coarse-to-fine registration for the position encoding. GARF~\cite{chng2022gaussian} provides a Gaussian based activation functions on the coarse-to-fine registration for more robust camera pose refinement. SCNeRF~\cite{jeong2021self} builds geometric loss optimization on the ray intersection re-projection error. Subsequent works~\cite{bosssamurai,kuang2022neroic,zhang2021ners} also incorperate the photometric loss from silhouette or mask, but requires accurate foreground segmentation. However, most of these approaches still depends on dense input views, which will not be effective for sparse scenarios. 

Different from these previous approaches, our approach explores the view-consistent constrains on the explicit surface geometry of neural surface representation, which provides more effective cues than rendered depth~\cite{truong2022sparf} to jointly learn neural surfance and refine camera poses in an end-to-end manner, without need any shape prior~\cite{zhang2021ners} or RGB-D input~\cite{sucar2021imap,azinovic2022neural,zhu2022nice}.

\input{method}

\input{exp}

\input{conclusion}


{\small
\bibliographystyle{ieee_fullname}
\bibliography{ref}
}

\end{document}


\title{Supplementary Materials of SC-NeuS: Consistent Neural Surface Reconstruction from Sparse and Noisy Views }


\maketitle


\input{quality}

\begin{table*}[htbp]
    \centering
    \caption{The quantitative comparing results from COLMAP, $w/o$ $L_{ncc}$ and ours approaches, evaluated on DTU dataset. }
    \resizebox{\textwidth}{!}{
    \begin{tabular}{|c|c|c|c|c|c|c|c|c|c|c|c|c|c|c|c|c|}
    \hline
        & \multicolumn{16}{c}{Translation \downarrow} \\ \hline
        Scan & 24 & 37 & 40 & 55 & 63 & 65 & 69 & 83 & 97 & 105 & 106 & 110 & 114 & 118 & 122 & mean \\ \hline
        COLMAP & 0.29  & 0.60  & 0.36  & 0.14  & \textbf{0.15}  & \textbf{0.11}  & \textbf{0.11}  & 0.25  & 0.38  & 0.38  & 0.12  & 0.15  & \textbf{0.05}  & \textbf{0.08}  & \textbf{0.16}  & 0.22  \\ 
        $w/o$ $L_{ncc}$ & 0.19  & 0.56  & \textbf{0.01}  & 0.43  & 0.38  & 0.13  & 0.12  & 0.21  & \textbf{0.21}  & 0.18  & \textbf{0.06}  & 0.20  & 0.15  & 0.16  & 0.44  & 0.23  \\ 
        Ours & \textbf{0.15}  & \textbf{0.23}  & 0.16  & \textbf{0.07}  & 0.16  & 0.17  & 0.16  & \textbf{0.07}  & 0.31  & \textbf{0.01}  & 0.17  & \textbf{0.12}  & 0.23  & 0.12  & 0.17  & \textbf{0.15}  \\ \hline
        & \multicolumn{16}{c}{Rotation (^{\circ}) \downarrow} \\ \hline
        Scan & 24 & 37 & 40 & 55 & 63 & 65 & 69 & 83 & 97 & 105 & 106 & 110 & 114 & 118 & 122 & mean \\ \hline
        COLMAP & 0.36  & 0.55  & 0.25  & 0.20  & 0.21  & 0.26  & 0.27  & 0.55  & 0.36  & 0.21  & 0.15  & 0.37  & 0.20  & 0.15  & 0.27  & 0.29  \\ 
        $w/o$ $L_{ncc}$ & 0.11  & 0.83  & \textbf{0.06}  & 0.30  & 0.23  & 0.70  & 0.11  & 0.45  & \textbf{0.14}  & 0.17  & 0.24  & 0.33  & 0.22  & 0.26  & 0.29  & 0.30  \\ 
        Ours & \textbf{0.07}  & \textbf{0.17}  & \textbf{0.06}  & \textbf{0.08}  & \textbf{0.06}  & \textbf{0.21}  & \textbf{0.10}  & \textbf{0.17}  & 0.21  & \textbf{0.06}  & \textbf{0.06}  & \textbf{0.18}  & \textbf{0.09}  & \textbf{0.08}  & \textbf{0.14}  & \textbf{0.12}  \\ \hline
        & \multicolumn{16}{c}{Chamfer Distance \downarrow} \\ \hline
        Scan & 24 & 37 & 40 & 55 & 63 & 65 & 69 & 83 & 97 & 105 & 106 & 110 & 114 & 118 & 122 & mean \\ \hline
        COLMAP & 10.76  & 9.53  & 2.66  & 1.29  & 1.77  & 2.73  & 1.32  & 9.16  & 6.92  & 1.64  & 1.79  & 13.40  & 0.61  & 1.37  & 1.88  & 4.46  \\ 
        $w/o$ $L_{ncc}$ & 1.75  & \textbf{1.51}  & 9.38  & \textbf{0.53}  & \textbf{1.16}  & \textbf{1.49}  & 1.98  & 6.54  & 2.17  & 1.50  & 1.71  & 9.20  & 1.58  & 9.59  & 1.63  & 3.45  \\ 
        Ours & \textbf{1.07}  & 2.14  & \textbf{1.55}  & 1.38  & 1.31  & 2.03  & \textbf{0.81}  & \textbf{2.95}  & \textbf{1.02}  & \textbf{1.39}  & \textbf{1.30}  & \textbf{1.62}  & \textbf{0.37}  & \textbf{0.88}  & \textbf{1.37}  & \textbf{1.41} \\ \hline
    \end{tabular}}
    \label{tab:colmap}
\end{table*}

\section{Comparison with COLMAP}
We additionally compare our approach with the state-of-the-art SfM system, COLMAP~\cite{schonberger2016structure}, \ssh{in sparse view scenario}.

Table~\ref{tab:colmap} demonstrates the quantitative \ssh{comparing} results on the \ssh{RMSE} accuracy of camera pose estimation and surface reconstruction. We can find that our approach outperforms the COLMAP in challenging sparse view setting, which is benefit from multi-view consistent constraint. 
Besides, we also show the quantitative results of $w/o$ $L_{ncc}$, a variant of our system that only uses view-consistent re-projection loss.
We can see that when only view-consistent re-projection loss, $w/o$ $L_{ncc}$ still achieves the comparable accuracy of camera estimation to COLMAP. When using full view-consistent constraints, the view-consistent patch-warping loss further improves the accuracy of camera pose estimation and then improves the surface reconstruction quality (see Ours in Table~\ref{tab:colmap}). 

\begin{figure}[ht]
    \centering
    \includegraphics[scale=0.6]{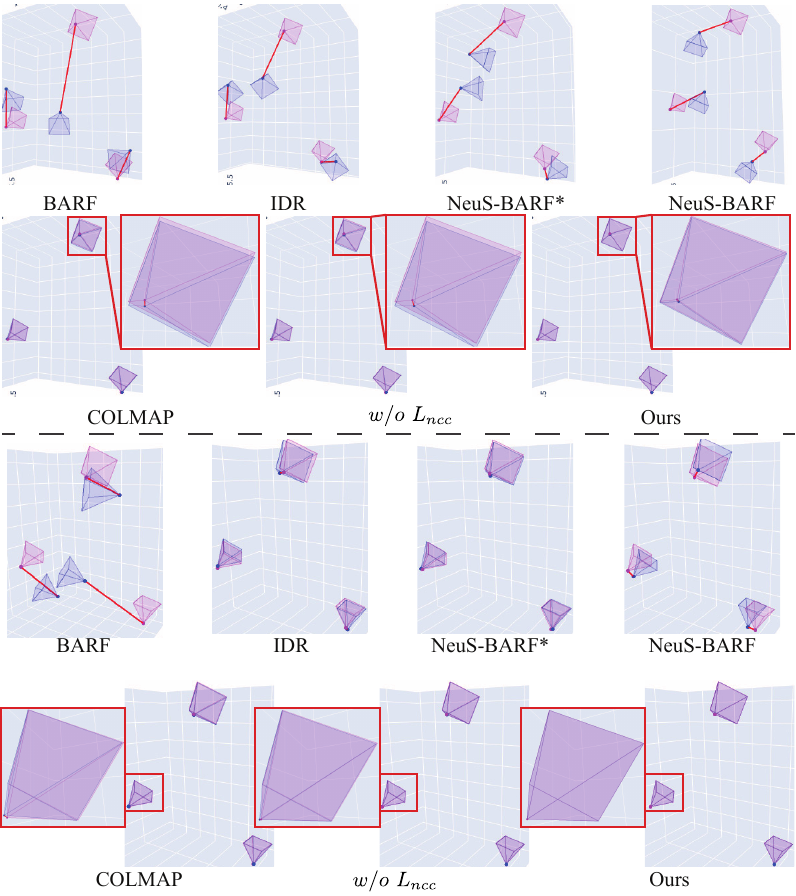}
    \caption{Two visual comparing results of camera poses estimation on scan37 (up) and scan105 (bottom) from DTU dataset using different comparing approaches. We show the optimized camera poses (blue) aligned to ground-truth camera poses (pink) with translation error (red line).}
    \label{fig:camera}
\end{figure}

Fig.~\ref{fig:camera} illustrates two visual comparing results for the camera poses estimation from different approaches on DTU dataset. It shows that only COLMAP, $w/o$ $L_{ncc}$ and Ours successfully converge for camera pose optimization, while the others fails to estimate accurate camera poses. Benefiting from view-consistent path-warping constraint, ours achieve more accurate camera poses estimation than COLMAP and $w/o$ $L_{ncc}$ (see close-up view in the bottom row in Fig.~\ref{fig:camera}).

\section{More Qualitative Results}
\begin{figure*}[ht]
    \centering
    \includegraphics[scale=1.0]{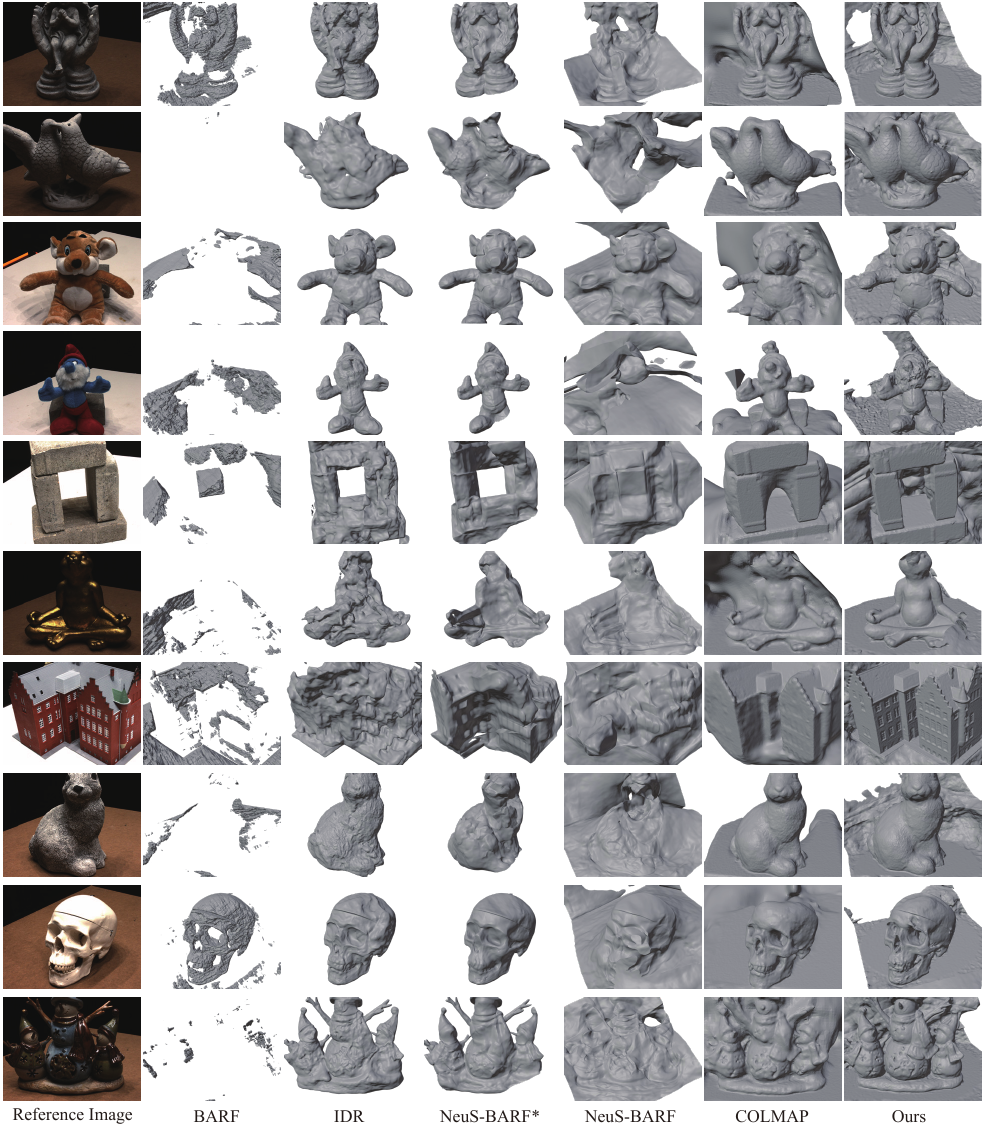}
    \caption{More visual comparison results on DTU dataset using different comparing approaches. All the results are the final converged results at the same training epochs from different comparing approaches.  }
    \label{fig:dtu_supp}
\end{figure*}

We present more qualitative comparing results of on DTU in Fig.~\ref{fig:dtu_supp} using different comparing approaches, including BARF, IDR, NeuS-BARF, NeuS-BARF*, COLMAP and Ours. Most of BARF's results fail to reconstruct complete surfaces. 
Although COLMAP achieves better surface reconstruction results than other previous approaches,  but still with limited on geometric detail reconstruction. 
On contrast, our approach achieves the satisfactory surface reconstruction results with fine-grained details, due to the effective multi-view consistent constraint for more accurate joint learning of neural surface and  camera pose estimation.

\begin{figure*}[ht]
    \centering
    \includegraphics[scale=0.8]{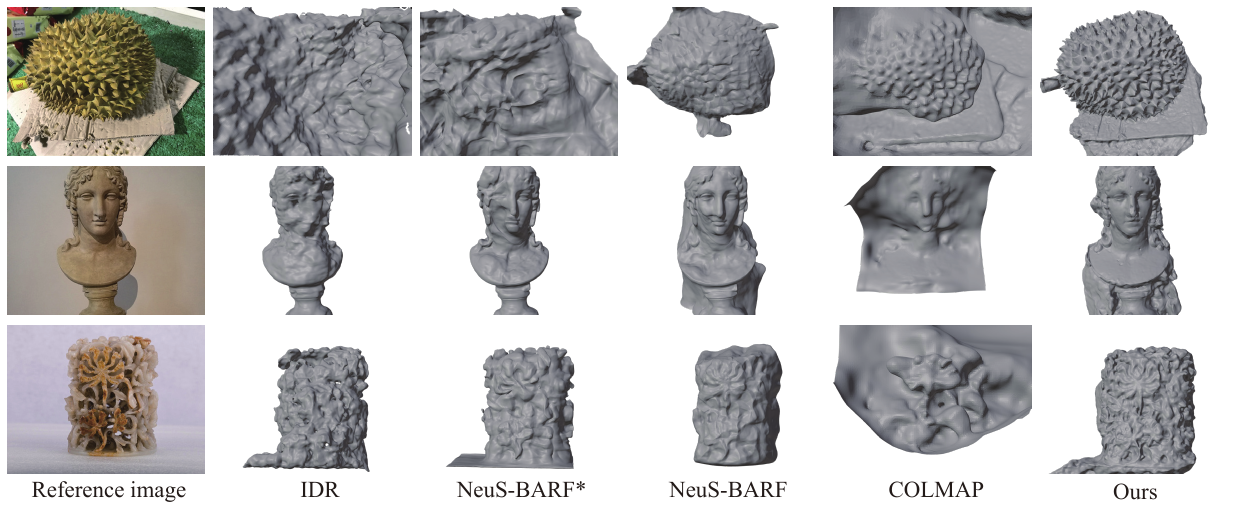}
    \caption{More visual comparison results on BlendedMVS dataset using different comparing approaches. All the results are the final converged results at the same training epochs from different comparing approaches.}
    \label{fig:bmvs_supp}
\end{figure*}

Besides, we also present more qualitative results on the BlendedMVS dataset in Fig.~\ref{fig:bmvs_supp}, including IDR, NeuS-BARF, NeuS-BARF*, COLMAP and ours. IDR, NeuS-BARF, NeuS-BARF* and COLMAP fail to reconstruct high quality surface on these challenging scenes, where input scenes' geometry is more complex surface than those from DTU dataset. 
Meanwhile, our approach still successfully estimate accurate camera poses with consistently better surface reconstruction quality. 


\section{More Intermediate Results}
\begin{figure*}[ht]
    \centering
    \includegraphics[scale=0.7]{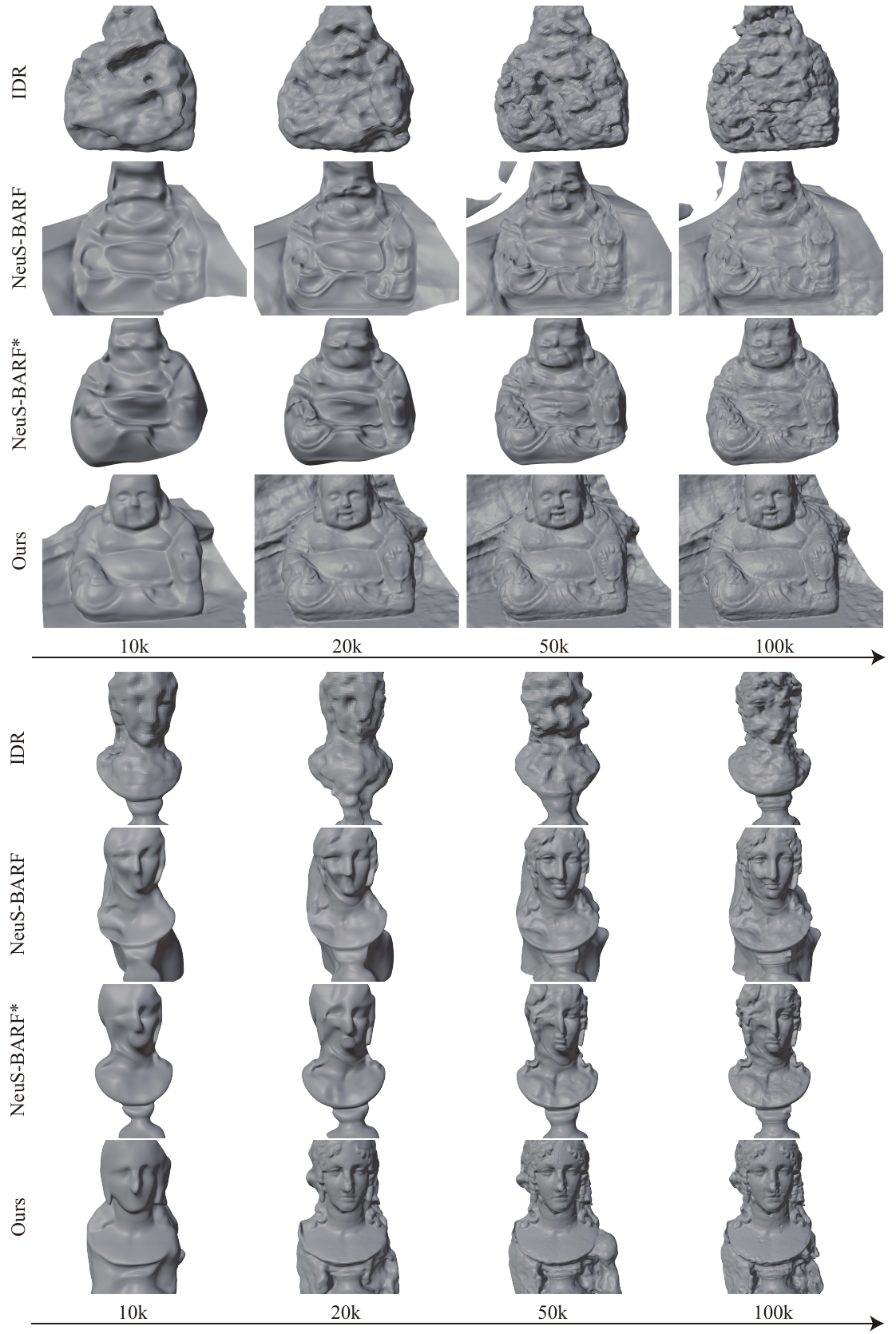}
    \caption{Intermediate results of different comapring approaches on DTU dataset (upper rows) and BlendedMVS dataset (bottom rows), including IDR, NeuS-BARF, NeuS-BARF* and ours respectively.}
    \label{fig:inter}
\end{figure*}

To make a comprehensive understanding of the joint learning process of our approach, we show more intermediate results of the neural surface  during the joint learning of neural surface and camera pose estimation. For comparison, we also show the intermediate results of IDR, NeuS-BARF and NeuS-BARF* at the same training epoch. Fig.~\ref{fig:inter} shows two visual comparing of the intermediate results of neural surface learning using different comparing approaches. It can be seen that our approach can converge faster with much better surface reconstruction quality than the other comparing approaches.  

\section{Time Analysis}
Our approach takes on average 4.5 hours to train an object on the platform of NVIDIA
RTX3090 GPU, and costs on average 8G GPU memory storage during the joint learning. We believe that our approach can be further accelerated using the state-of-the-art framework, such as Instant-NGP~\cite{muller2022instant}, for much faster time efficiency.

\section{Comparison with Geo-NeuS-BARF}
\begin{figure*}[h]
\begin{center}
   \includegraphics[width=\linewidth]{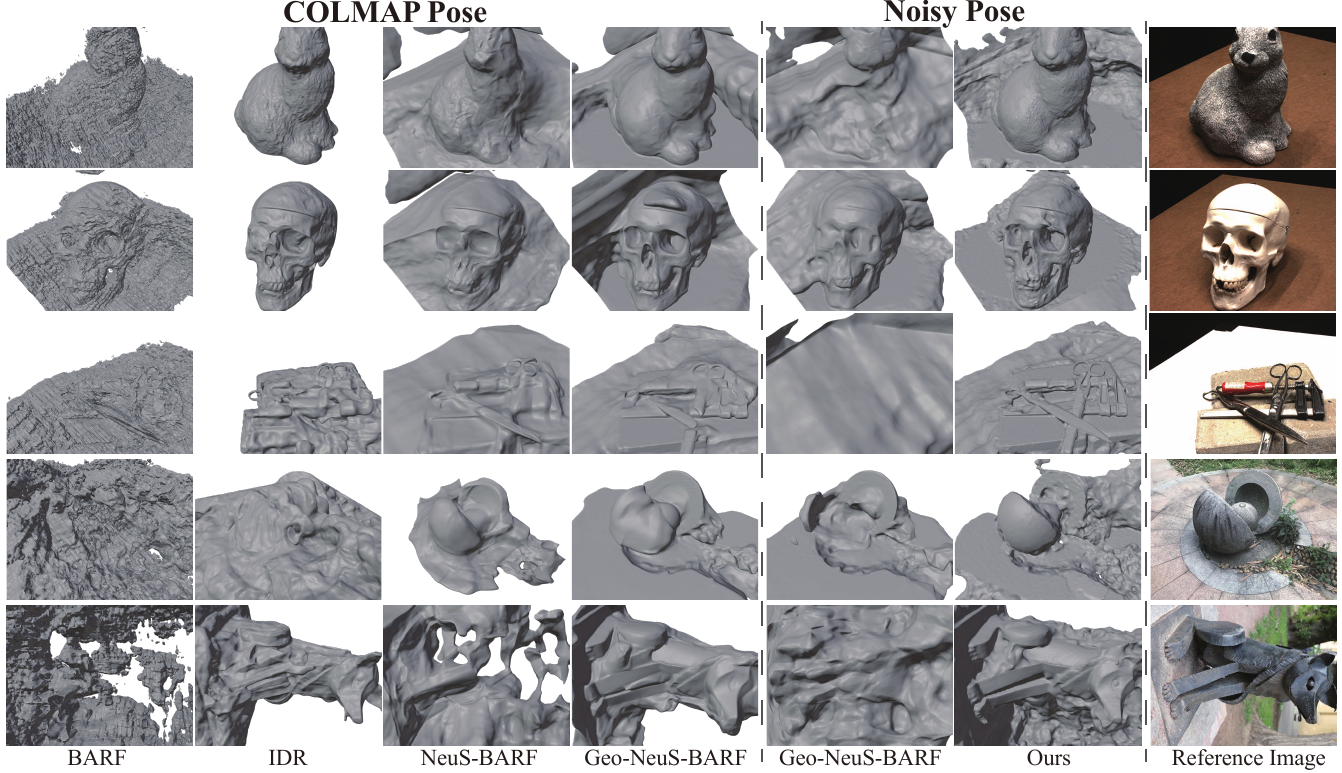}
\end{center}
   \caption{The visual comparison results for different approaches with camera poses initialized from COLMAP or noise on DTU dataset (top three) and BlendedMVS (bottom two) dataset.}
\label{fig:result}
\end{figure*}

We implemented a BARF-based version of Geo-NeuS\cite{fu2022geo}\footnote{https://github.com/GhiXu/Geo-Neus}, and made comparison with ours. 
Table~\ref{tab:exp} (bottom rows) shows the reconstruction accuracy for Geo-NeuS-BARF and ours (given the same noisy input). Our approach achieves much better reconstruction quality than  Geo-NeuS-BARF, which means that our regularization from explicit geometry of neural surface and view-consistent constraints takes effects.  Fig.~\ref{fig:result} also shows some visual comparison results, where ours can achieve better surface reconstruction with details preserved. More comparison results will be included in our revision.

\begin{table}[ht]
    \centering
    \caption{The RMSE 
    and CD 
    accuracy comparison with camera poses initialized from COLMAP or noise on DTU dataset.}
    \resizebox{0.4\textwidth}{!}{
    \begin{tabular}{c|ccccc}
    \hline
        & Initial & Trans $\downarrow$ & Rot ($^{\circ}$) $\downarrow$ & CD $\downarrow$ \\ \hline
        BARF & COLMAP & 0.41 & 0.35 & 4.36 \\
        IDR & COLMAP & 1.09 & 0.53 & 4.55 \\
        NeuS & COLMAP & 1.42 & 0.46 &  4.57 \\
        Geo-NeuS & COLMAP & 0.32 & 0.33 & 2.21 \\
        \hline
        Geo-NeuS & noisy & 8.60 & 15.42 & 5.00 \\
        Ours & noisy & \textbf{0.15} & \textbf{0.12} & \textbf{1.41} \\ \hline
    \end{tabular}\label{tab:exp}
    }
\end{table}

\section{Comparison with initialization using COLMAP}
We have made experiments with initialization using COLMAP for previous approaches.
Table~\ref{tab:exp} (upper rows) shows the reconstruction accuracy of previous approaches evaluated on DTU dataset. Similarly, our approach (even with noisy initialization) also achieves better reconstruction quality than previous approaches with COLMAP initialization.  Fig.~\ref{fig:result} shows some visual comparison results. 

\begin{table}[ht]
    \centering
    \caption{The RMSE 
    and CD 
    accuracy comparison with camera poses initialized from COLMAP or noise on DTU dataset.}
    \resizebox{0.4\textwidth}{!}{
    \begin{tabular}{c|ccccc}
    \hline
        & Initial & Trans $\downarrow$ & Rot ($^{\circ}$) $\downarrow$ & CD $\downarrow$ \\ \hline
        BARF & COLMAP & 0.41 & 0.35 & 4.36 \\
        IDR & COLMAP & 1.09 & 0.53 & 4.55 \\
        NeuS & COLMAP & 1.42 & 0.46 &  4.57 \\
        Geo-NeuS & COLMAP & 0.32 & 0.33 & 2.21 \\
        \hline
        Geo-NeuS & noisy & 8.60 & 15.42 & 5.00 \\
        Ours & noisy & \textbf{0.15} & \textbf{0.12} & \textbf{1.41} \\ \hline
    \end{tabular}\label{tab:exp}
    }
\end{table}

\section{Reconstruction quality V.S. 2D-Match quality}
\begin{figure}[h]
\begin{center}
   \includegraphics[width=0.95\linewidth]{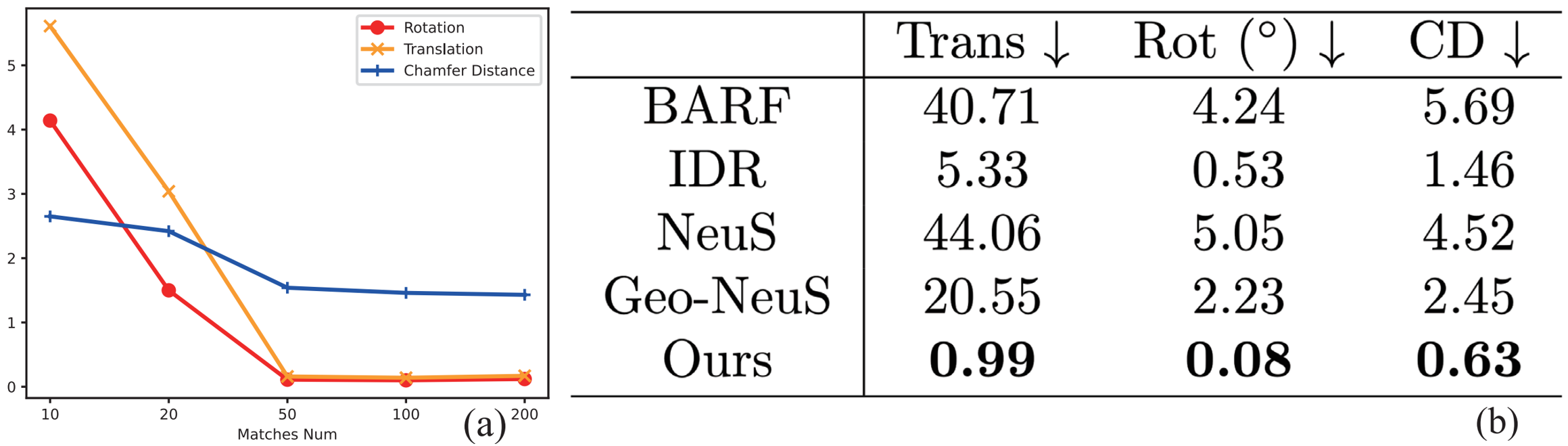}
\end{center}
   \caption{(a) The accuracy curves v.s. 2D-match number in our approach. (b) The  accuracy comparison for dense and noisy views.}
\label{fig:match}
\end{figure}

As we have discussed in Sec. 4.5, our approach will be influenced by the 2D-match quality, which is similar for feature-based approaches like COLMAP. 
We have made evaluation on our reconstruction quality v.s. 2D-match quality. 
Fig.~\ref{fig:match} (a) shows the accuracy curves of camera pose estimation and surface reconstruction v.s. 2D-match number. 
\vspace{-0.3mm}

\section{About extension to dense views} 
Our approach can be easily extended to \emph{dense} and \emph{noisy} camera views. We have made evaluation on dense and noisy view scenarios of our approach according to your suggestion. Fig.~\ref{fig:match} (b) shows the reconstruction accuracy comparison for different approaches from DTU dataset. Our approach can also achieve much better reconstruction accuracy than previous approaches. 


{\small
\bibliographystyle{ieee_fullname}
\bibliography{ref}
}

%% file: method.tex
\section{SC-NeuS}\label{sec_method}
Given sparse view images (as few as 3) with noisy camera poses of an object, we aim at reconstructing the surface represented by neural implicit function and jointly optimizing the camera poses. Specifically, for sparse input views $I=\{I_i\}$ with noisy camera poses $T=\{T_i\}$ ($i \in \{1,2,3\}$), we adopt to represent the object's geometry as signed distance field (SDF) $f(x,\bm{\theta})$ ($x \in R^3$, $\bm{\theta}$ is the MLP parameter), and render its appearance using volume rendering from an extra radiance filed $c(x,\bm{\theta_c},\bm{v})$ as provided by NeuS~\cite{wang2021neus}. 

By introducing effective multi-view constraints across sparse views, we 
propose an new joint learning strategy, called SC-NeuS, for both signed distance field $f(x,\bm{\theta})$ learning and camera poses $T=\{T_i\}$ optimization. Fig.~\ref{system} demonstrates the main pipeline of our SC-NeuS framework in an end-to-end learning manner.


\textbf{From Multi-view Constraints to Geometry-consistent Surface Learning.} Unlike the previous approaches~\cite{chng2022gaussian,lin2021barf,meng2021gnerf,yariv2020multiview} that perform the joint deep implicit geometry learning and camera pose optimization using photometric loss across dense input views \emph{independently}, we adopt to exploit multi-view constraints as extra effective regularization to constraint the surface learning. Due to the bias gap between the volume rendering integral and point-based SDF modeling~\cite{fugeo}, instead of relying on multi-view depth rendering prior from the neural surface to multi-view depth priors~\cite{chen2021mvsnerf,truong2022sparf,fugeo}, we propose to utilize more multi-view regularizations directly from the explicit surface geometry of the neural surface for a better multi-view surface reconstruction. Our key observation is that the geometry cues (points or patches) locating on the shape surface should be consistently observed across multi-views, which is intuitively an effective constraints for geometry-consistent surface learning, especially in sparse scenarios.

Specifically, we first derive an fast \emph{differentiable} point intersection on the explicit surface of signed distance filed $f(x,\bm{\theta})$ (Sec.~\ref{sec:intersection}). Then we provide view-consistent losses for two kinds of on-surface geometry cues (3D sparse points and patches) based on our differentiable point intersection, including view-consistent re-projection loss and patch-warping loss (Sec.~\ref{sec:loss}), to effectively regularize the joint learning of signed distance field $f(x,\bm{\theta})$ and camera poses $T$ . Since the intersection derived by our approach is differentiable for both the neural surface parameters $\bm{\theta}$ and camera poses $T$, our neural surface learning can be performed in an end-to-end manner without any other supervisions.

\begin{figure}[t]
\begin{center}
   \includegraphics[width=0.97\linewidth]{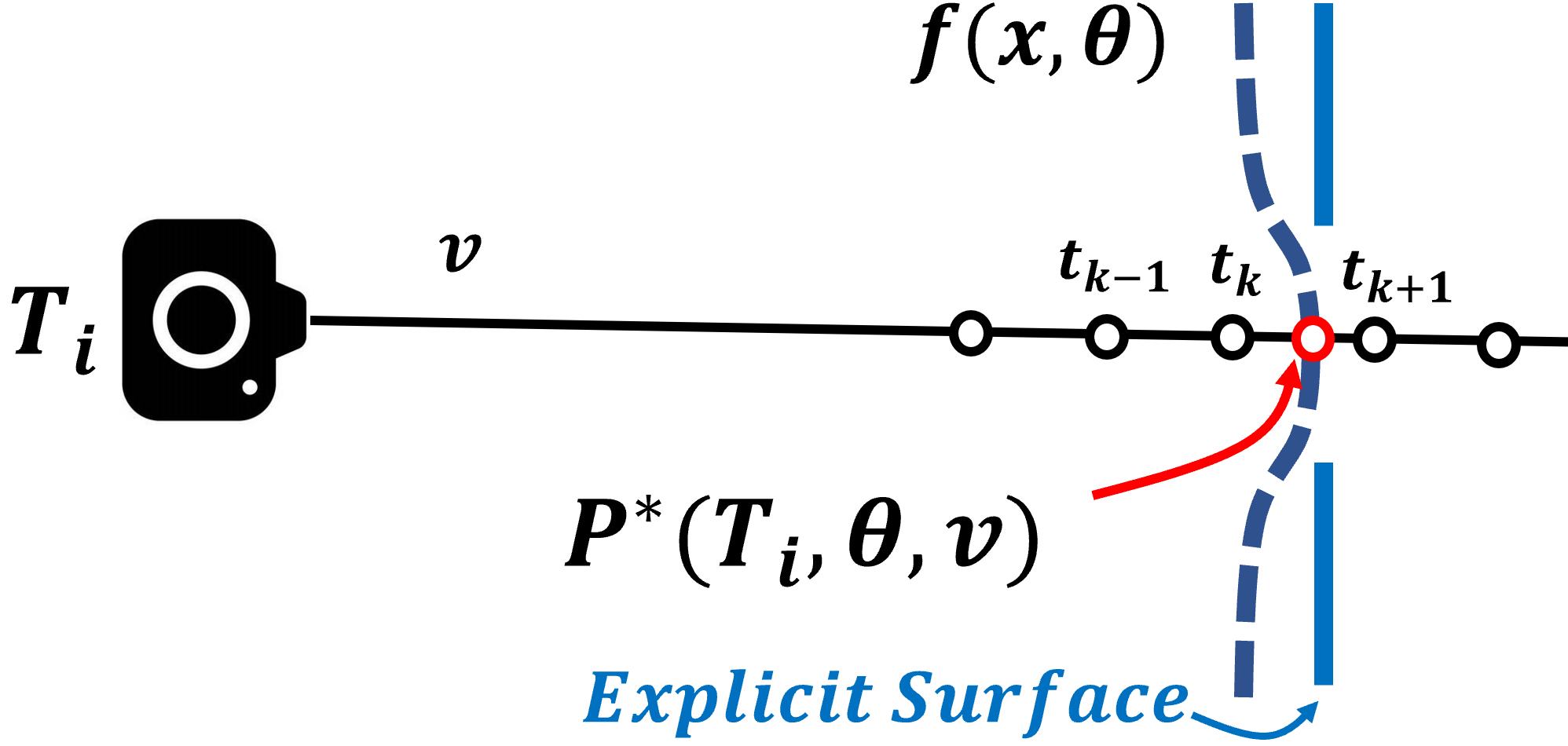}
\end{center}
   \caption{The illustration of our fast differentiable on-surface intersection  $\bm{P}^{*}(T_i,\bm{\theta},\bm{v})$, between the explicit surface of signed distance filed $f(x,\bm{\theta})$ and camera view $T_i$ along the casting ray $\bm{v}$.}
\label{multi_view}
\end{figure}

\subsection{Differentiable On-surface Intersection}\label{sec:intersection}

To enable multi-view consistent constraints, the essential requirement of the geometry cues is that they need \emph{locate} on the explicit surface, i.e., the zero level set of the signed distance field $f(x,\bm{\theta})$. Considering a 2D feature point $p \in R^2$ in the reference image $I_i$ with camera pose $T_i$, we seek to compute its intersection point $\bm{P}^{*} \in R^3$ on the surface geometry of signed distance field $f(x,\bm{\theta})$. According to volume rendering of the signed distance function~\cite{mildenhall2021nerf,wang2021neus}, there exists a ray length value $t^{*}$ such that:
\[
\bm{P}^{*} = \bm{c_i} + t^{*} \bm{v}, \quad f(\bm{P}^{*},\bm{\theta}) = 0,
\]
where $\bm{c_i}$ and $\bm{v}$ are the camera center point and casting ray of $p$ respectively.

Although IDR~\cite{yariv2020multiview} have provided a differentiable intersection derivation for $\bm{P}^{*}$, however, which is somewhat too slow to enable an efficient neural surface learning. Therefore, we propose a new differentiable on-surface intersection for fast neural surface learning. Specifically, as shown in Fig.~\ref{multi_view}, we first uniformly sample points in the casting ray $\bm{v}$ of 2D feature point $p$ with sampling depth value set $\mathbf{T}=\{t_k\}$. Then we find the depth value $t_k$ such that $f(\bm{c_i}+t_k \bm{v},\bm{\theta}) f(\bm{c_i}+t_{k+1} \bm{v},\bm{\theta}) < 0$. Finally, we move $t_k$ along the casting ray $\bm{v}$ to the on-surface intersection $\bm{P}^{*}(T_i,\bm{\theta},\bm{v})$ following:

\begin{equation}
    \bm{P}^{*}(T_i,\bm{\theta},\bm{v}) = \bm{c_i}+t_k \bm{v} - \frac{\bm{v}}{\langle \frac{\partial f}{\partial x}, \bm{v} \rangle} f(\bm{c_i}+t_k \bm{v}, \bm{\theta}).
\end{equation}


\begin{figure}[t]
\begin{center}
   \includegraphics[width=0.94\linewidth]{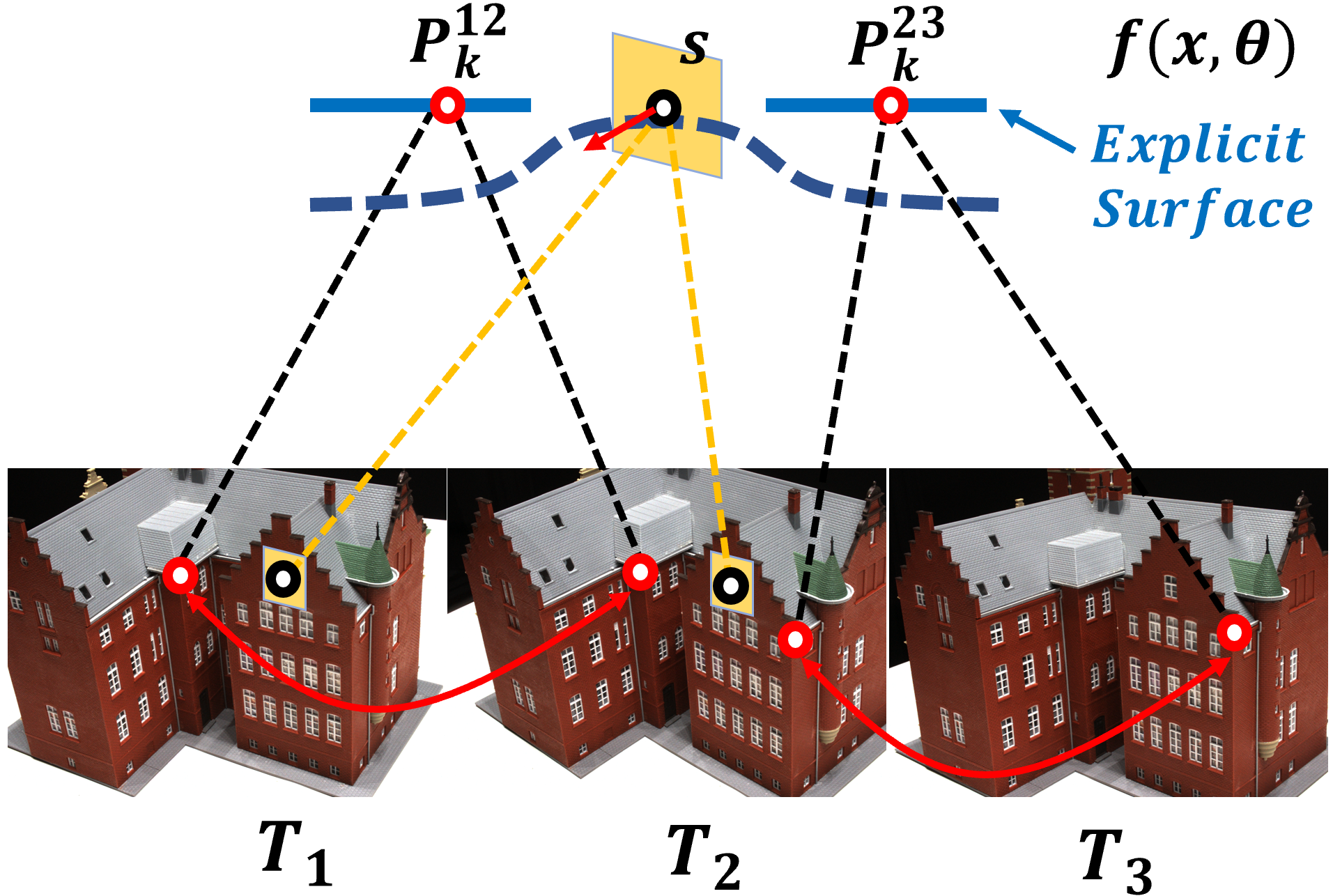}
\end{center}
   \caption{An example illustration of view-consistent losses defined on two kinds of on-surface geometry cues, i.e., 3D sparse points (colored red) and patches (colored orange), for a sparse view (3 views) input case from DTU dataset.}
\label{loss}
\end{figure}

\subsection{View-Consistent Loss}\label{sec:loss}

Based on our differentiable intersection, we further define effective losses to neural surface learning in the multi-view scenario. Specifically, we utilize two kinds of on-surface geometry cues, i.e., 3D sparse points and patches (Fig.~\ref{loss}), and formulate  view-consistent losses for these on-surface geometry cues, including view-consistent re-projection loss and view-consistent patch-warping loss respectively.

\textbf{View-consistent Re-projection Loss.} Considering a pair of 2D feature correspondence ($p_k^i,p_k^j$) from reference image $I_i$ (camera pose $T_i$) and target image $I_j$  (camera pose $T_j$) with $p_k^i \in I_i, p_k^j \in I_j$, we compute the on-surface intersection 3D point $\bm{P}_k^{ij}$ via our differentiable intersection. By re-projecting $\bm{P}_k^{ij}$ back to $I_i$ and $I_j$, we get the re-projection location as $\bar{p}_k^i = \pi(\bm{P}_k^{ij} , T_i)$, $\bar{p}_k^j = \pi(\bm{P}_k^{ij} , T_j)$, where $\pi(\cdot)$ is the camera projection operator. For a geometry-consistent surface reconstruction, the re-projection error between $p_k^i \to \bar{p}_k^i$ and $p_k^j \to \bar{p}_k^j$ should be minimized. Then we formulate the view-consistent re-projection loss $L_r$ for all of possible sparse correspondence  as:
\[
L_r = \sum_{i,j} \sum_{k \in N_k} (|p_k^i-\pi(\bm{P}_k^{ij} , T_i)| + ||p_k^j-\pi(\bm{P}_k^{ij} , T_j)|).
\]

\textbf{View-consistent Patch-warping Loss.} We also consider the on-surface patch (Fig.~\ref{loss}) to utilize the geometric structure constraints to further improve the neural surface learning. Similar to the patch warping in traditional MVS method~\cite{labatut2007efficient,schonberger2016pixelwise,schonberger2016structure}, we warp the on-surface patch to multi-view images but in a differentiable way using our differentiable multi-view intersection. Specifically, for a small patch $\bm{s}$ on the surface which is observed by image pair $I_i,I_j$, we represent the plane equation of $\bm{s}$ in the camera coordinate of the reference image $I_i$ as:
\[
\bm{n}^T \bm{p} + d = 0,
\]
where $\bm{p}(T_i,T_j)$ is the differentiable multi-view intersection point from $I_i,I_j$ with camera poses $T_i,T_j$, $\bm{n}$ is the normal computed with automatic differentiation of the signed distance filed $f(x,\bm{\theta})$ at $\bm{p}(T_i,T_j)$. Suppose that the $\bm{s}$ is projected to $I_i,I_j$ to obtain image patches $\bm{s}_i \in I_i, \bm{s}_j \in I_j$ respectively, for image pixel $x \in \bm{s}_i$ and its corresponding pixel $x' \in \bm{s}_j$, we have:
\[
x = \bm{H}x' , \\ \bm{H} = K_i(R_iR_j^t - \frac{R_i(R_i^T t_i - R_j^T t_j) \bm{n}^T }{d})K_j^{-1},
\]
where $\bm{H}$ is the homography matrix, $T_i=\{R_i|t_i\},T_j=\{R_j|t_j\}$, $K_i,K_j$ are the intrinsic camera matrix for image pair $I_i,I_j$.

We use the normalization cross correlation (NCC) of patches ($\bm{s}_i,\bm{s}_j$) as the view-consistent patch-warping loss as :
\[
L_{ncc}(\bm{s}_i,\bm{s}_j) = \frac{Cov(I_i(\bm{s}_i),I_j(\bm{s}_j))}{Var(I_i(\bm{s}_i))Var(I_j(\bm{s}_j))},
\]
where $Cov$ and $Var$ donates the covariance and variance for color identity of patches ($\bm{s}_i,\bm{s}_j$) respectively.

\subsection{Training Strategy}
Based on the view-consistent losses, we formulate the objective function $E$ as:
\begin{equation}
    E = L_{color} + \lambda_r L_r + \lambda_{ncc} L_{ncc} + \lambda_{reg} L_{reg},
\end{equation}
with $L_r$ and $L_{ncc}$ are the view-consistent re-projection loss and patch-warping loss defined above, and $L_{color}$ and $L_{reg}$ are the color rendering loss and Eikonal regularization loss proposed by NeuS~\cite{wang2021neus} as:
\[
L_{color} = \frac{1}{N} \sum_{i}^{N}|\mathcal{R}(f(x,\bm{\theta}),c(x,\bm{\theta_c},\bm{v}),T_i) - I_i|,
\]
\[
L_{reg} = \frac{1}{M} \sum_i^{M}(||\bigtriangledown f_{\theta}||_2 -1)^2,
\]
where $\mathcal{R}(f(x,\bm{\theta}),c(x,\bm{\theta_c},\bm{v}),T_i)$ is the volume rendering image from $f(x,\bm{\theta}),c(x,\bm{\theta_c},\bm{v})$ to view $T_i$.

So in summary, we propose to jointly learn the signed distance filed $f(x,\bm{\theta})$, radian field $c(x,\bm{\theta_c},\bm{v})$ and camera poses $T=\{T_i\}$ to optimize the objective function $E$ in an end-to-end manner following:
\begin{equation}
    \{\bm{\theta^{*}},\bm{\theta_c^{*}},T^{*}\} = \arg\min_{\bm{\theta},\bm{\theta_c},T} E.
    \label{main}
\end{equation}


\textbf{Network Training.} In the early stage during the network training, since the signed distance field $f(x,\bm{\theta})$ doesn't converge very well, we choose a warm-up strategy to assist the convergence. Specifically, for the differentiable on-surface intersection of 3D sparse point, we use the rendering depth information of the recent signed distance field $f(x,\bm{\theta})$ to filter out outlier on-surface point intersection. Given a 2D feature point $p$ and its on-surface intersection point $\bm{P}^{*}$ along casting ray $\bm{v}$, we compute its depth value $t_{d}$~\cite{truong2022sparf} and get its 3D point re-projection $\bm{P}_d = \bm{c}_i + t_d \bm{v}$. If the distance between $\bm{P}^{*}$ and $\bm{P}_d$ is larger than a threshold, we set $\bm{P}^{*}=\bm{P}_d$ to perform the joint learning. After the warm-up for a certain training epoch, we conduct the learning according to equation~\ref{main} until the final convergence for both signed distance field $f(x,\bm{\theta})$ and camera poses $T$. 

\textbf{Coarse-to-Fine Learning.} As like BARF~\cite{lin2021barf}, we also adopt the similar coarse-to-fine positional encoding learning strategy, for a better convergence during the joint learning of neural surface and camera poses. Please refer to BARF~\cite{lin2021barf} for more details.

%% file: exp.tex
\section{Experiments and Analysis}\label{sec_exp}

To evaluate the effectiveness of our SC-NeuS, we conduct surface reconstruction experiments from sparse and noisy views of public dataset, by comparing with previous state-of-the-art approaches. 
Thereafter, we also give an ablation study and analysis of the main components in our approach to make a comprehensive understanding for our SC-NeuS.

\textbf{Implementation details.}
We adopt the similar architecture of IDR~\cite{yariv2020multiview} and NeuS~\cite{wang2021neus} by using a MLP (8 hidden layers with hidden width of 256) for SDF (Signed Distance Function) $f$ and another MLP (4 hidden layers with hidden width of 256) for radiance filed $c$. 
For 2D feature correspondence, we use the out-of-the-box key-point detection and extraction model, ASLFeat~\cite{DBLP:conf/cvpr/LuoZBCZ0LFQ20}, and key-point feature matching model, SuperGlue~\cite{DBLP:conf/cvpr/SarlinDMR20}.
We randomly sample 512 rays and select 256 2D correspondences per batch and train our model for 100K iterations on a single NVIDIA RTX3090 GPU.

\begin{table*}[ht]
    \centering
    \caption{The quantitative comparison results in terms of RMSE accuracy (both translation and rotation errors) of camera pose estimation from different comparing approaches on DTU dataset. "Noisy Input" means the initial RMSE accuracy of the input noisy camera poses v.s. the grounth truth camera poses. "NeuS-BARF*" represents NeuS-BARF that ultilizes extra object mask supervision.}
    \resizebox{\textwidth}{!}{
    \begin{tabular}{c|c|c|c|c|c|c|c|c|c|c|c|c|c|c|c|c}
    \hline
    & \multicolumn{16}{c}{Translation $\downarrow$} \\ \hline
    Scan & 24 & 37 & 40 & 55 & 63 & 65 & 69 & 83 & 97 & 105 & 106 & 110 & 114 & 118 & 122 & Mean \\ \hline
    Noisy Input & 28.03  & 43.79  & 27.47  & 47.88  & 20.64  & 10.17  & 11.94  & 57.06  & 39.78  & 39.85  & 50.93  & 47.04  & 30.19  & 6.87  & 28.79 & 32.70 \\  \hline
    BARF & 42.30  & 56.48  & 17.08  & 46.44  & 13.11  & 0.72  & 21.69  & 48.69  & 3.33  & 48.34  & 44.98  & 46.16  & 22.76  & 1.03  & 13.22  & 28.42  \\ 
    IDR & 41.87  & 41.76  & 43.42  & 42.67  & 1.12  & 0.39  & 0.82  & 1.12  & 10.15  & 2.95  & 59.49  & 47.74  & 46.70  & 0.62  & 27.13  & 24.53  \\ 
    NeuS-BARF & 30.73  & 39.47  & 27.13  & 10.32  & 24.20  & 3.47  & 23.84  & 53.24  & 13.83  & 6.72  & 38.74  & 14.03  & 0.34  & 31.71  & 30.44  & 23.21  \\ 
    NeuS-BARF$^{*}$ & 45.31  & 39.86  & 41.69  & 43.91  & 1.16  & 0.48  & 1.28  & 1.76  & 1.73  & 1.44  & 53.33  & 46.98  & 0.17  & 0.66  & 29.56  & 20.62  \\ 
    Ours & \textbf{0.15} & \textbf{0.23} & \textbf{0.16}  & \textbf{0.07}  & \textbf{0.16} & \textbf{0.17}  & \textbf{0.16}  & \textbf{0.07}  & \textbf{0.31}  & \textbf{0.01}  & \textbf{0.17}  & \textbf{0.12}  & \textbf{0.23}  & \textbf{0.12}  & \textbf{0.17}  & \textbf{0.15}  \\ \hline
    \hline
    & \multicolumn{16}{c}{Rotation ($^{\circ}$) $\downarrow$} \\ \hline
    Scan & 24 & 37 & 40 & 55 & 63 & 65 & 69 & 83 & 97 & 105 & 106 & 110 & 114 & 118 & 122 & Mean \\ \hline
    Noisy Input & 22.41  & 14.32  & 19.94  & 18.90  & 14.55  & 21.26  & 15.62  & 18.12  & 15.46  & 15.46  & 27.55  & 18.64  & 24.41  & 10.81  & 21.17 & 18.57  \\ \hline
    BARF & 21.42  & 19.45  & 18.36  & 19.38  & 25.64  & 3.60  & 15.15  & 32.07  & 1.28  & 14.91  & 23.36  & 19.38  & 29.89  & 1.07  & 22.82  & 17.85  \\ 
    IDR & 27.37  & 33.45  & 24.82  & 21.59  & 4.02  & 0.55  & 0.67  & 1.66  & 2.67  & 0.52  & 51.26  & 21.09  & 22.25  & 2.70  & 45.99  & 17.37  \\ 
    NeuS-BARF & 21.79  & 25.76  & 17.97  & 15.59  & 11.05  & 3.20  & 14.97  & 23.43  & 5.28  & 4.69  & 42.43  & 10.61  & 0.27  & 15.78  & 12.43  & 15.02  \\ 
    NeuS-BARF$^{*}$ & 32.95  & 26.01  & 32.69  & 16.62  & 1.93  & 1.45  & 0.22  & 0.55  & 0.59  & 0.19  & 26.77  & 16.28  & 0.27  & 0.55  & 18.83  & 11.73  \\ 
    Ours & \textbf{0.07} & \textbf{0.17}  & \textbf{0.06}  & \textbf{0.08}  & \textbf{0.06} & \textbf{0.21}  & \textbf{0.10}  & \textbf{0.17}  & \textbf{0.21}  & \textbf{0.06}  & \textbf{0.06}  & \textbf{0.18}  & \textbf{0.09}  & \textbf{0.08}  & \textbf{0.14}  & \textbf{0.12}  \\ \hline
    \end{tabular}
    }
    \label{tab:pose}
\end{table*}

\begin{table*}[htbp]
    \centering
    \caption{The quantitative comparison results of Chamfer Distance accuracy for the surface reconstruction using different comparing approaches on DTU dataset.}
    \resizebox{\textwidth}{!}{
        \begin{tabular}{c|c|c|c|c|c|c|c|c|c|c|c|c|c|c|c|c}
    \hline
        Scan & 24 & 37 & 40 & 55 & 63 & 65 & 69 & 83 & 97 & 105 & 106 & 110 & 114 & 118 & 122 & Mean \\ \hline
        BARF & 7.62  & 8.82  & 8.86  & 8.34  & 8.19  & 6.85  & 8.58  & 9.44  & 7.87  & 9.93  & 7.01  & 7.87  & 7.96  & 6.32  & 7.83  & 8.10  \\ 
        IDR & 8.48  & 9.21  & 8.74  & 10.40  & 11.44  & 7.19  & 3.47  & 7.32  & 9.47  & 4.67  & 8.12  & 9.21  & 7.71  & 7.21  & 8.72  & 8.09  \\ 
        NeuS-BARF & 7.93  & 7.89  & 8.51  & 9.39  & 9.07  & 8.53  & 8.23  & 8.78  & 9.77  & 8.31  & 9.33  & 6.35  & 8.88  & 7.59  & 9.68  & 8.55  \\ 
        NeuS-BARF$^{*}$ & 9.22  & 9.90  & 8.31  & 9.16  & 9.60  & 5.94  & 3.75  & 7.04  & 5.64  & 2.37  & 8.59  & 9.00  & 1.10  & 3.47  & 8.67  & 6.78  \\
        Ours & \textbf{1.07}  & \textbf{2.14}  & \textbf{1.55}  & \textbf{1.38}  & \textbf{1.31}  & \textbf{2.03}  & \textbf{0.81}  & \textbf{2.95}  & \textbf{1.02}  & \textbf{1.39}  & \textbf{1.30}  & \textbf{1.62}  & \textbf{0.37}  & \textbf{0.88}  & \textbf{1.37}  & \textbf{1.41} \\ \hline
    \end{tabular}\label{tab:dtu_mesh}
    }
\end{table*}

\subsection{Experimental Settings}
\textbf{Dataset.}
Similar with previous neural surface reconstruction approaches~\cite{jeong2021self,yariv2020multiview,wang2021neus}, we choose to evaluate our approach on the public DTU dataset~\cite{DBLP:journals/ijcv/AanaesJVTD16} with 15 different object scan. The DTU dataset contains from 49 to 64 images at a resolution of $1200 \times 1600$ for each object scan with known camera intrinsic matrix and ground truth camera poses. For sparse views, 
we follow \cite{long2022sparseneus} and ~\cite{lin2021barf} to randomly select as few as 3 views for each object scan, and then synthetically perturb its camera pose 
with an additive Gaussian noise $\mathcal{N}(0, 0.15)$, thus collecting a sparse version of DTU dataset for the subsequent evaluation. Besides, we also evaluate on 7 challenging scenes from low-res set of the BlendedMVS dataset~\cite{DBLP:conf/cvpr/0008LLZRZFQ20} which includes $31-143$ images at a resolution of $768 \times 576$. Similar to pre-processing as like our DTU dataset, we also select 3 views from them and obtain their noisy initial poses.

\textbf{Baselines.}
We compare our approach with previous state-of-the-art approaches which also perform neural surface reconstruction by joint learning of neural surface and camera poses, including BARF~\cite{lin2021barf}and IDR~\cite{yariv2020multiview}. Besides, although NeuS~\cite{wang2021neus} doesn't conduct camera pose optimization,  since it is a state-of-the-art neural surface reconstruction approach, we also compare our approach with NeuS by incorporating the coarse-to-fine strategy of BARF~\cite{lin2021barf}, called ``NeuS-BARF'', to enable a fair comparison. 
Since IDR use an extra object mask for neural surface learning, for fair comparison of IDR and NeuS-BARF, we additionally conduct an experiment by using extra object mask supervision for NeuS-BARF, named ``NeuS-BARF*'', during the subsequent evaluations.


\begin{figure*}[htbp]
    \centering
    \includegraphics[scale=0.9]{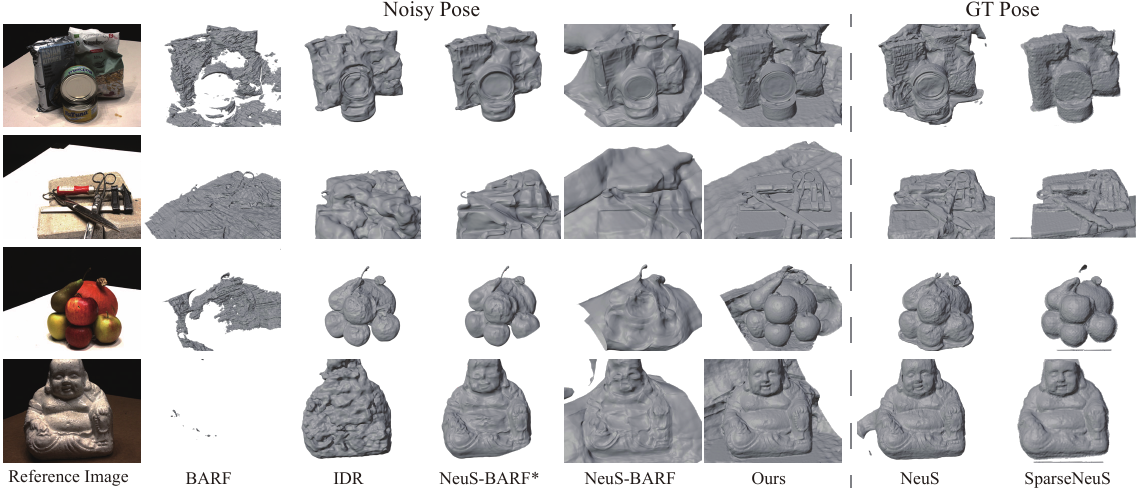}
    \caption{The visual comparing surface reconstruction results using different neural surface reconstruction approaches, with noisy input camera poses (left columns) and ground truth input camera poses (right columns) on DTU dataset.}
    \label{fig:dtu}
\end{figure*}

\subsection{Evaluation on DTU Dataset}

\textbf{Camera Pose Comparison.} Table~\ref{tab:pose} demonstrates the average RMSE accuracy (including both translation and rotation error) between the estimated camera poses and the ground truth camera poses on DTU dataset, using different comparing approaches, including BARF, IDR, NeuS-BARF,  NeuS-BARF* and ours respectively. Among all the comparing approaches, the NeRF-like approach BARF, achieves worse RMSE accuracy than the other approaches. This makes sense since other approaches (including our approach) adopt the signed distance field to represent the object's geometry, which is more powerful than the radiance filed used in BARF.
Although IDR and NeuS-BARF (NeuS-BARF*) achieve various RMSE accuracy in each object scan of DTU dataset respectively, in average they achieve the same level of RMSE accuracy, which means they perform similar camera pose estimation quality. 

In contrast, our approach significantly outperforms all the other baselines in the RMSE accuracy (in both the translation and rotation errors) for camera pose estimation. This shows the multi-view consistent constraints used in our SC-NeuS takes effects for camera pose estimation, than the other baseline approaches which performs the camera pose optimization along with the neural surface learning with single-view independent regularization.


\textbf{Surface Reconstruction Quality.} We also compare the surface reconstruction quality between the different comparing approaches.  Table~\ref{tab:dtu_mesh} demonstrates the quantitative results on Chamfer Distance metric using different approaches evaluated on DTU dataset. Similarly, our approach achieve consistently much better Chamfer Distance accuracy than the other comparing approaches. 
Fig.~\ref{fig:dtu} illustrates some visual comparison results of the comparing approaches.
Even though BARF can achieve acceptable camera pose estimation quality, it still fails to achieve fine surface reconstruction results (see the first row of Fig.~\ref{fig:dtu}). 

Besides, due to lack of extra mask object supervision, NeuS-BARF can't achieve accurate enough camera pose estimation and thus fails to reconstruct fine object surface. This demonstrates that coarse-to-fine position embedding proposed in BARF is not effective to sparse view setting, even utilizing the neural signed distance filed representation as like NeuS. In contrast, our approach choose view-consistent constraints to regularize the joint learning of neural surface representation and camera pose, leading to geometry-consistent surface reconstruction with fine-grained details. Please see the fine-grained detail reconstruction by our approach, which is also better than the state-of-the-art neural surface reconstruction approach like NeuS and SparseNeuS, even with ground truth camera poses as input (Fig.~\ref{fig:dtu}). Please refer to our supplementary materials for more comparing results.


\begin{figure*}[ht]
    \centering
    \includegraphics[scale=0.85]{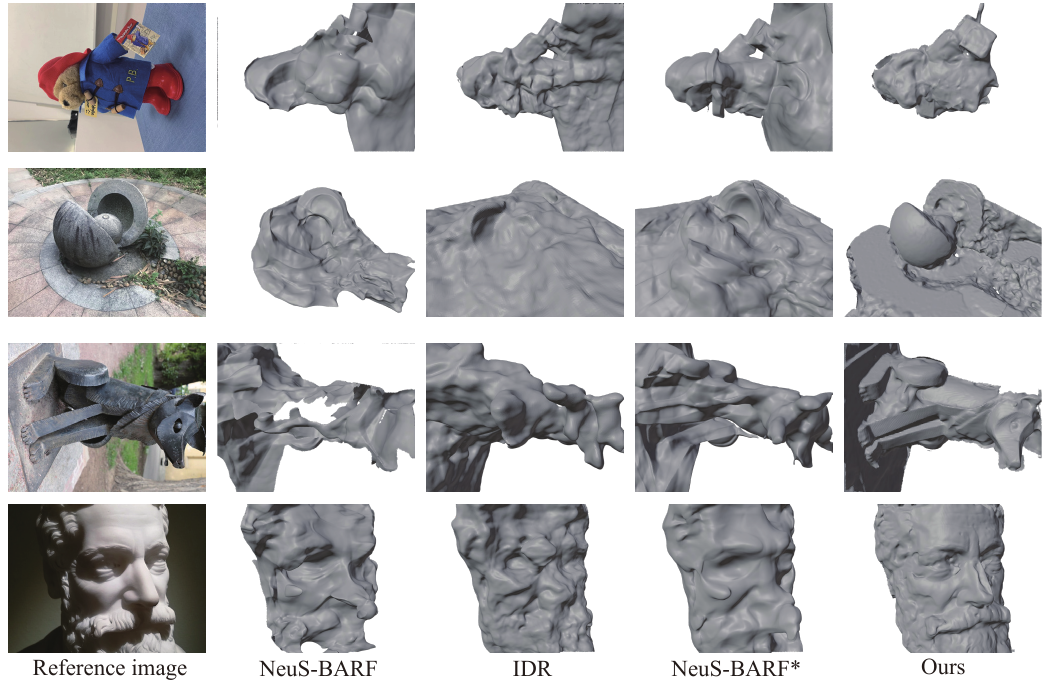}
    \caption{The visual comparing surface reconstruction results using different neural surface reconstruction approaches, including NeuS-BARF, NeuS-BARF*, IDR and our approach, with sparse view noisy input camera poses on the BlendedMVS dataset.}
    \label{fig:bmvs}
\end{figure*}

%% file: conclusion.tex
\subsection{Evaluation on BlendedMVS Dataset}

Except from the DTU dataset, we also perform evaluation on BlendedMVS dataset to see how our approach behave across different kinds of datasets. 
Fig.~\ref{fig:bmvs} shows some visual comparing surface reconstruction results using different comparing approaches, including NeuS-BARF, NeuS-BARF*, IDR and our approach. According to the comparison, our approach can achieve much better surface reconstruction quality with fine-grained details than the other approaches. Here we don't include BARF for visual comparison, since BARF fails to converge in most of the comparing cases. Please refer to our supplementary materials for more quantitative and qualitative comparing results using different approaches on BlendedMVS dataset. 

\subsection{Ablation and Analysis}
The re-projection loss and patch-warping loss serve as two main components of our approach.  We conduct an ablation study experiment, to see how these two losses take effect on the final quality of both surface reconstruction and camera pose estimation. 
\begin{figure}[ht]
    \centering
    \includegraphics[scale=0.5]{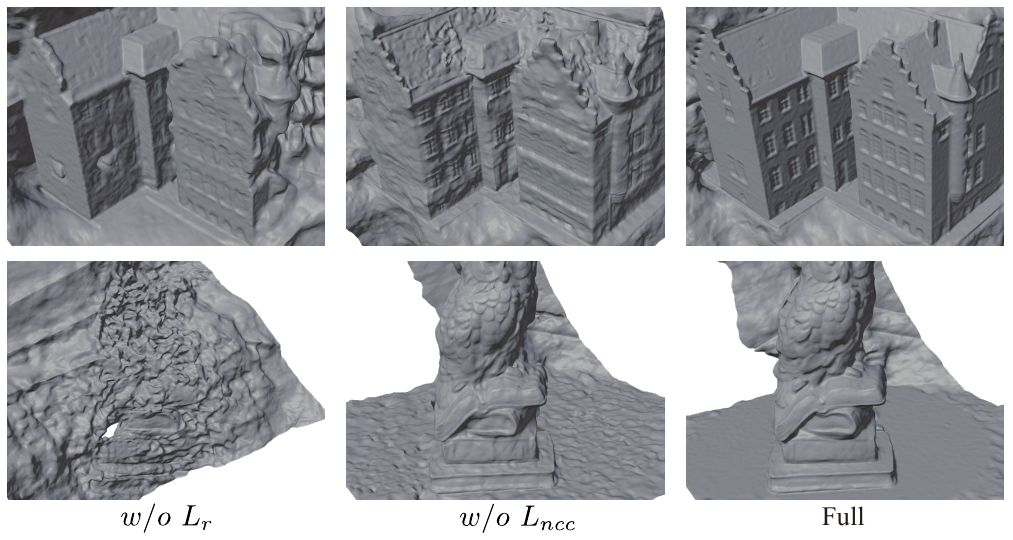}
    \caption{The visual results by different variants of our system.}
    \label{fig:ab}
\end{figure}

\begin{table}[ht]
    \centering
    \caption{The RMSE (both translation and rotation) and Chamfer Distance (CD) accuracy by different variants of our system.}
    \begin{tabular}{c|ccc}
    \hline
          & Translation $\downarrow$ & Rotation ($^{\circ}$) $\downarrow$ & CD $\downarrow$ \\ \hline
        w/o $L_r$ & 22.71 & 11.77 & 6.44 \\
        w/o $L_{ncc}$ & 0.23 & 0.30 & 3.44 \\
        Full & 0.15 & 0.12 & 1.41 \\ \hline
    \end{tabular}\label{tab:ab}
\end{table}

\textbf{View-consistent Re-projection.} We first implement a variant version of our full system without using the view-consistent re-projection loss, termed as 'w/o $L_r$', and perform the surface reconstruction on the DTU dataset. Table~\ref{tab:ab} shows the average RMSE accuracy for camera pose estimation quality and CD accuracy for the surface reconstruction quality for 'w/o $L_r$' and our full system (termed as 'Full'). We can see there are large accuracy decrease for both RMSE and CD between 'w/o $L_r$' and 'Full' systems. This means the view-consistent re-projection loss serves major contribution in our SC-NeuS for the final geometry-consistent surface reconstruction and accurate camera pose estimation. But please note that 'w/o $L_r$' still outperforms other comparing approaches including BARF, IDR and NeuS-BAFR, by achieving better average RMSE accuracy and CD accuracy in Table~\ref{tab:pose} and ~\ref{tab:dtu_mesh}. 

\textbf{View-consistent Patch-warping.} We also implement a variant system without using the view-consistent patch-warping loss, termed as 'w/o $L_{ncc}$'. According to the average RMSE accuracy and CD accuracy comparison between 'w/o $L_{ncc}$' and 'Full' in Table~\ref{tab:ab}, we can see that 'w/o $L_{ncc}$' also achieve worse accuracy values than 'Full' in both RMSE for camera pose estimation and CD for surface reconstruction quality, even thougth the quality decreases are not that much compared with those from 'w/o $L_r$' to 'Full'. 

Fig.~\ref{fig:ab} shows the visual comparing surface reconstruction results  of two example from DTU dataset, using 'w/o $L_r$', 'w/o $L_{ncc}$' and 'Full' respectively. We can see there are certain surface quality decrease for our full system ('Full') without using the view-consistent re-projection loss ('w/o $L_r$'). Even though our approach can achieve fine surface reconstruction without using view-consistent patch-warping loss (see the results of 'w/o $L_{ncc}$'), we can obvious fine-grained details enhancement  by adding the view-consistent loss to our full system (see the results of 'Full'). This means that view-consistent patch-warping loss takes more effective for fine-grained details, while   view-consistent re-projection loss works better to boost up the joint learning quality of neural surface and camera pose.


\subsection{Limitation and Discussion}
Our approach's first limitation is that influence from the quality of 2D feature point's matching. Without enough feature point matching in challenging cases like low texture or light changing, our approach couldn't perform well for nice surface reconstruction results. Large camera poses variation between sparse views would also make our approach failed for feasible joint optimization. In the further, we would like to use more robust explicit surface priors for high reliable neural surface reconstruction. 

\section{Conclusion}\label{sec_conclusion}
Joint learning for the neural surface representation and camera pose remains to be a challenging problem, especially for sparse scenarios. This paper propose a new joint learning strategy, called SC-NeuS, which explores multi-view constraints directly from the explicit geometry of the neural surface. Compared with previous neural surface reconstruction approaches, our SC-NeuS achieves consistently better surface reconstruction quality and camera pose estimation accuracy, for geometry-consistent neural surface reconstruction results with fine-grained details. We hope that our approach can inspire more efforts to the neural surface reconstruction from sparse view images, to enable more feasible real-world applications in this community.   

%% file: quality.tex
\section{Clarify for the Reconstruction Quality Evaluation}
\begin{figure}[t]
    \centering
    \includegraphics[scale=0.8]{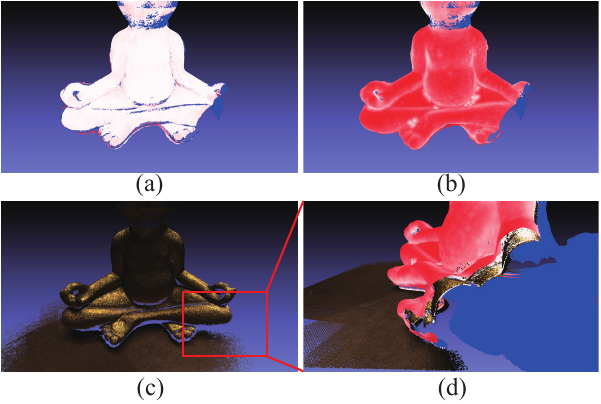}
    \caption{An \ssh{tiny} illustration of \ssh{reconstructed mesh alignment quality to the ground-truth point cloud } 
    \ssh{with (a) or without (b)} our ICP registration refinement, evaluated by Chamfer Distance with white (a) representing lower CD and red (b) representing high CD error. 
    \ssh{The original Sim3 transformation alignment to the ground-truth point cloud} (c) \ssh{obviously leads to bad quality alignment (d) for unfair high CD accuracy,} 
    while our ICP registration refinement achieves better surface alignment with lower CD accuracy. } 
    \label{fig:align}
\end{figure}

\ssh{When performing the surface reconstruction quality evaluation,  previous neural surface reconstruction approaches~\cite{wang2021neus,yariv2020multiview,fugeo,darmon2022improving} often adopt the}
official evaluation script by DTU dataset~\cite{DBLP:journals/ijcv/AanaesJVTD16} to compute the  Chamfer Distance accuracy, \ssh{where the reconstructed meshes are aligned to the ground-truth point cloud using the Sim3 transformation, which is computed from the alignment of estimated camera poses to the ground-truth camera poses}. However, 
\ssh{such Sim3 transformation would be easily } 
affected by the camera pose estimation accuracy, \ssh{thus leading to unfair comparison between the individual reconstructed surface quality respectively.} 
For example, the reconstructed mesh sometimes \ssh{couldn't} be well aligned to ground-truth point cloud using the Sim3 transformation from estimated camera poses to ground-truth poses  (Fig.~\ref{fig:align} (c-d) , with a higher Chamfer Distance results (Fig.~\ref{fig:align} (b)), \ssh{but the reconstructed mesh could be very similar with the ground-truth point cloud}. 

In order to 
\ssh{make a fair comparison on the surface reconstruction quality itself,}
we \ssh{propose to further align the reconstructed mesh to the ground-truth point cloud, by } additionally applying an ICP registration to refine the original \ssh{Sim3} transformation \ssh{based alignment} using the Open3D~\cite{Zhou2018} tools. \ssh{Using such further refinement,}  the reconstructed mesh can be better aligned to ground-truth point cloud
obtaining lower Chamfer Distance (Fig.~\ref{fig:align} (a)), which is better to reflect the surface reconstruction quality itself. 
\ssh{Considering that the} ICP registration \ssh{would also fail without good camera pose initialization, }
\ssh{in such failure cases we still use the original Sim3 transformation alignment to compute the Chamfer Distance accuracy.} 

\ssh{So that's how we get the Chamfer Distance accuracy in Table 2 of our main paper, showing a fair comparison on the surface reconstruction accuracy using different comparing approaches on the DTU dataset~\cite{DBLP:journals/ijcv/AanaesJVTD16}. We can see that our approach still achieves consistently better Chamfer Distance accuracy than the other comparing approaches, only for the surface reconstruction quality itself.}

%% file: main.bbl
\begin{thebibliography}{10}\itemsep=-1pt

\bibitem{DBLP:journals/ijcv/AanaesJVTD16}
Henrik Aan{\ae}s, Rasmus~Ramsb{\o}l Jensen, George Vogiatzis, Engin Tola, and
  Anders~Bjorholm Dahl.
\newblock Large-scale data for multiple-view stereopsis.
\newblock {\em IJCV.}, 120(2):153--168, 2016.

\bibitem{atzmon2020sal}
Matan Atzmon and Yaron Lipman.
\newblock Sal: Sign agnostic learning of shapes from raw data.
\newblock In {\em IEEE CVPR}, pages 2565--2574, 2020.

\bibitem{azinovic2022neural}
Dejan Azinovi{\'c}, Ricardo Martin-Brualla, Dan~B Goldman, Matthias
  Nie{\ss}ner, and Justus Thies.
\newblock Neural rgb-d surface reconstruction.
\newblock In {\em IEEE CVPR}, pages 6290--6301, 2022.

\bibitem{barron2022mip}
Jonathan~T Barron, Ben Mildenhall, Dor Verbin, Pratul~P Srinivasan, and Peter
  Hedman.
\newblock Mip-nerf 360: Unbounded anti-aliased neural radiance fields.
\newblock In {\em IEEE CVPR}, pages 5470--5479, 2022.

\bibitem{bosssamurai}
Mark Boss, Andreas Engelhardt, Abhishek Kar, Yuanzhen Li, Deqing Sun,
  Jonathan~T Barron, Hendrik Lensch, and Varun Jampani.
\newblock Samurai: Shape and material from unconstrained real-world arbitrary
  image collections.
\newblock In {\em Advances in Neural Information Processing Systems}, 2022.

\bibitem{chen2021mvsnerf}
Anpei Chen, Zexiang Xu, Fuqiang Zhao, Xiaoshuai Zhang, Fanbo Xiang, Jingyi Yu,
  and Hao Su.
\newblock Mvsnerf: Fast generalizable radiance field reconstruction from
  multi-view stereo.
\newblock In {\em IEEE CVPR}, pages 14124--14133, 2021.

\bibitem{chng2022gaussian}
Shin-Fang Chng, Sameera Ramasinghe, Jamie Sherrah, and Simon Lucey.
\newblock Gaussian activated neural radiance fields for high fidelity
  reconstruction and pose estimation.
\newblock In {\em ECCV}, pages 264--280. Springer, 2022.

\bibitem{darmon2022improving}
Fran{\c{c}}ois Darmon, B{\'e}n{\'e}dicte Bascle, Jean-Cl{\'e}ment Devaux,
  Pascal Monasse, and Mathieu Aubry.
\newblock Improving neural implicit surfaces geometry with patch warping.
\newblock In {\em IEEE CVPR}, pages 6260--6269, 2022.

\bibitem{deng2022depth}
Kangle Deng, Andrew Liu, Jun-Yan Zhu, and Deva Ramanan.
\newblock Depth-supervised nerf: Fewer views and faster training for free.
\newblock In {\em IEEE CVPR}, pages 12882--12891, 2022.

\bibitem{fugeo}
Qiancheng Fu, Qingshan Xu, Yew-Soon Ong, and Wenbing Tao.
\newblock Geo-neus: Geometry-consistent neural implicit surfaces learning for
  multi-view reconstruction.
\newblock In {\em Advances in Neural Information Processing Systems}.

\bibitem{jensen2014large}
Rasmus Jensen, Anders Dahl, George Vogiatzis, Engin Tola, and Henrik Aan{\ae}s.
\newblock Large scale multi-view stereopsis evaluation.
\newblock In {\em IEEE CVPR}, pages 406--413, 2014.

\bibitem{jeong2021self}
Yoonwoo Jeong, Seokjun Ahn, Christopher Choy, Anima Anandkumar, Minsu Cho, and
  Jaesik Park.
\newblock Self-calibrating neural radiance fields.
\newblock In {\em IEEE CVPR}, pages 5846--5854, 2021.

\bibitem{jiang2020local}
Chiyu Jiang, Avneesh Sud, Ameesh Makadia, Jingwei Huang, Matthias Nie{\ss}ner,
  Thomas Funkhouser, et~al.
\newblock Local implicit grid representations for 3d scenes.
\newblock In {\em CVPR}, pages 6001--6010, 2020.

\bibitem{kar2017learning}
Abhishek Kar, Christian H{\"a}ne, and Jitendra Malik.
\newblock Learning a multi-view stereo machine.
\newblock {\em Advances in neural information processing systems}, 30, 2017.

\bibitem{kim2022infonerf}
Mijeong Kim, Seonguk Seo, and Bohyung Han.
\newblock Infonerf: Ray entropy minimization for few-shot neural volume
  rendering.
\newblock In {\em IEEE CVPR}, pages 12912--12921, 2022.

\bibitem{kuang2022neroic}
Zhengfei Kuang, Kyle Olszewski, Menglei Chai, Zeng Huang, Panos Achlioptas, and
  Sergey Tulyakov.
\newblock Neroic: neural rendering of objects from online image collections.
\newblock {\em ACM Transactions on Graphics (TOG)}, 41(4):1--12, 2022.

\bibitem{labatut2007efficient}
Patrick Labatut, Jean-Philippe Pons, and Renaud Keriven.
\newblock Efficient multi-view reconstruction of large-scale scenes using
  interest points, delaunay triangulation and graph cuts.
\newblock In {\em IEEE ICCV}, pages 1--8. IEEE, 2007.

\bibitem{lin2021barf}
Chen-Hsuan Lin, Wei-Chiu Ma, Antonio Torralba, and Simon Lucey.
\newblock Barf: Bundle-adjusting neural radiance fields.
\newblock In {\em IEEE CVPR}, pages 5741--5751, 2021.

\bibitem{long2022sparseneus}
Xiaoxiao Long, Cheng Lin, Peng Wang, Taku Komura, and Wenping Wang.
\newblock Sparseneus: Fast generalizable neural surface reconstruction from
  sparse views.
\newblock In {\em ECCV}, pages 210--227. Springer, 2022.

\bibitem{DBLP:conf/cvpr/LuoZBCZ0LFQ20}
Zixin Luo, Lei Zhou, Xuyang Bai, Hongkai Chen, Jiahui Zhang, Yao Yao, Shiwei
  Li, Tian Fang, and Long Quan.
\newblock Aslfeat: Learning local features of accurate shape and localization.
\newblock In {\em {IEEE} {CVPR}}, pages 6588--6597, 2020.

\bibitem{meng2021gnerf}
Quan Meng, Anpei Chen, Haimin Luo, Minye Wu, Hao Su, Lan Xu, Xuming He, and
  Jingyi Yu.
\newblock Gnerf: Gan-based neural radiance field without posed camera.
\newblock In {\em IEEE ICCV}, pages 6351--6361, 2021.

\bibitem{mildenhall2021nerf}
Ben Mildenhall, Pratul~P Srinivasan, Matthew Tancik, Jonathan~T Barron, Ravi
  Ramamoorthi, and Ren Ng.
\newblock Nerf: Representing scenes as neural radiance fields for view
  synthesis.
\newblock {\em Communications of the ACM}, 65(1):99--106, 2021.

\bibitem{niemeyer2022regnerf}
Michael Niemeyer, Jonathan~T Barron, Ben Mildenhall, Mehdi~SM Sajjadi, Andreas
  Geiger, and Noha Radwan.
\newblock Regnerf: Regularizing neural radiance fields for view synthesis from
  sparse inputs.
\newblock In {\em IEEE CVPR}, pages 5480--5490, 2022.

\bibitem{oechsle2021unisurf}
Michael Oechsle, Songyou Peng, and Andreas Geiger.
\newblock Unisurf: Unifying neural implicit surfaces and radiance fields for
  multi-view reconstruction.
\newblock In {\em IEEE ICCV}, pages 5589--5599, 2021.

\bibitem{park2019deepsdf}
Jeong~Joon Park, Peter Florence, Julian Straub, Richard Newcombe, and Steven
  Lovegrove.
\newblock Deepsdf: Learning continuous signed distance functions for shape
  representation.
\newblock In {\em IEEE CVPR}, pages 165--174, 2019.

\bibitem{peng2020convolutional}
Songyou Peng, Michael Niemeyer, Lars Mescheder, Marc Pollefeys, and Andreas
  Geiger.
\newblock Convolutional occupancy networks.
\newblock In {\em ECCV}, pages 523--540. Springer, 2020.

\bibitem{roessle2022dense}
Barbara Roessle, Jonathan~T Barron, Ben Mildenhall, Pratul~P Srinivasan, and
  Matthias Nie{\ss}ner.
\newblock Dense depth priors for neural radiance fields from sparse input
  views.
\newblock In {\em IEEE CVPR}, pages 12892--12901, 2022.

\bibitem{DBLP:conf/cvpr/SarlinDMR20}
Paul{-}Edouard Sarlin, Daniel DeTone, Tomasz Malisiewicz, and Andrew
  Rabinovich.
\newblock Superglue: Learning feature matching with graph neural networks.
\newblock In {\em IEEE {CVPR}}, pages 4937--4946, 2020.

\bibitem{schonberger2016structure}
Johannes~L Schonberger and Jan-Michael Frahm.
\newblock Structure-from-motion revisited.
\newblock In {\em IEEE CVPR}, pages 4104--4113, 2016.

\bibitem{schonberger2016pixelwise}
Johannes~L Sch{\"o}nberger, Enliang Zheng, Jan-Michael Frahm, and Marc
  Pollefeys.
\newblock Pixelwise view selection for unstructured multi-view stereo.
\newblock In {\em ECCV}, pages 501--518. Springer, 2016.

\bibitem{snavely2006photo}
Noah Snavely, Steven~M Seitz, and Richard Szeliski.
\newblock Photo tourism: exploring photo collections in 3d.
\newblock In {\em ACM SIGGRAPH}, pages 835--846. 2006.

\bibitem{sucar2021imap}
Edgar Sucar, Shikun Liu, Joseph Ortiz, and Andrew~J Davison.
\newblock imap: Implicit mapping and positioning in real-time.
\newblock In {\em IEEE CVPR}, pages 6229--6238, 2021.

\bibitem{trevithick2021grf}
Alex Trevithick and Bo Yang.
\newblock Grf: Learning a general radiance field for 3d representation and
  rendering.
\newblock In {\em IEEE ICCV}, pages 15182--15192, 2021.

\bibitem{truong2022sparf}
Prune Truong, Marie-Julie Rakotosaona, Fabian Manhardt, and Federico Tombari.
\newblock Sparf: Neural radiance fields from sparse and noisy poses.
\newblock {\em arXiv e-prints}, pages arXiv--2211, 2022.

\bibitem{wang2021neus}
Peng Wang, Lingjie Liu, Yuan Liu, Christian Theobalt, Taku Komura, and Wenping
  Wang.
\newblock Neus: Learning neural implicit surfaces by volume rendering for
  multi-view reconstruction.
\newblock {\em Advances in Neural Information Processing Systems},
  34:27171--27183, 2021.

\bibitem{wang2021ibrnet}
Qianqian Wang, Zhicheng Wang, Kyle Genova, Pratul~P Srinivasan, Howard Zhou,
  Jonathan~T Barron, Ricardo Martin-Brualla, Noah Snavely, and Thomas
  Funkhouser.
\newblock Ibrnet: Learning multi-view image-based rendering.
\newblock In {\em IEEE CVPR}, pages 4690--4699, 2021.

\bibitem{wang2021nerfmm}
Zirui Wang, Shangzhe Wu, Weidi Xie, Min Chen, and Victor~Adrian Prisacariu.
\newblock Ne{RF}$--$: Neural radiance fields without known camera parameters.
\newblock {\em arXiv preprint arXiv:2102.07064}, 2021.

\bibitem{wei2021nerfingmvs}
Yi Wei, Shaohui Liu, Yongming Rao, Wang Zhao, Jiwen Lu, and Jie Zhou.
\newblock Nerfingmvs: Guided optimization of neural radiance fields for indoor
  multi-view stereo.
\newblock In {\em IEEE CVPR}, pages 5610--5619, 2021.

\bibitem{xu2019multi}
Qingshan Xu and Wenbing Tao.
\newblock Multi-scale geometric consistency guided multi-view stereo.
\newblock In {\em IEEE CVPR}, pages 5483--5492, 2019.

\bibitem{xu2020pvsnet}
Qingshan Xu and Wenbing Tao.
\newblock Pvsnet: Pixelwise visibility-aware multi-view stereo network.
\newblock {\em arXiv preprint arXiv:2007.07714}, 2020.

\bibitem{yao2018mvsnet}
Yao Yao, Zixin Luo, Shiwei Li, Tian Fang, and Long Quan.
\newblock Mvsnet: Depth inference for unstructured multi-view stereo.
\newblock In {\em ECCV}, pages 767--783, 2018.

\bibitem{DBLP:conf/cvpr/0008LLZRZFQ20}
Yao Yao, Zixin Luo, Shiwei Li, Jingyang Zhang, Yufan Ren, Lei Zhou, Tian Fang,
  and Long Quan.
\newblock Blendedmvs: {A} large-scale dataset for generalized multi-view stereo
  networks.
\newblock In {\em {IEEE} {CVPR}}, pages 1787--1796, 2020.

\bibitem{yariv2021volume}
Lior Yariv, Jiatao Gu, Yoni Kasten, and Yaron Lipman.
\newblock Volume rendering of neural implicit surfaces.
\newblock {\em Advances in Neural Information Processing Systems},
  34:4805--4815, 2021.

\bibitem{yariv2020multiview}
Lior Yariv, Yoni Kasten, Dror Moran, Meirav Galun, Matan Atzmon, Basri Ronen,
  and Yaron Lipman.
\newblock Multiview neural surface reconstruction by disentangling geometry and
  appearance.
\newblock {\em Advances in Neural Information Processing Systems},
  33:2492--2502, 2020.

\bibitem{yu2021pixelnerf}
Alex Yu, Vickie Ye, Matthew Tancik, and Angjoo Kanazawa.
\newblock pixelnerf: Neural radiance fields from one or few images.
\newblock In {\em IEEE CVPR}, pages 4578--4587, 2021.

\bibitem{zhang2021ners}
Jason Zhang, Gengshan Yang, Shubham Tulsiani, and Deva Ramanan.
\newblock Ners: neural reflectance surfaces for sparse-view 3d reconstruction
  in the wild.
\newblock {\em Advances in Neural Information Processing Systems},
  34:29835--29847, 2021.

\bibitem{zhang2022relpose}
Jason~Y Zhang, Deva Ramanan, and Shubham Tulsiani.
\newblock Relpose: Predicting probabilistic relative rotation for single
  objects in the wild.
\newblock In {\em ECCV}, pages 592--611. Springer, 2022.

\bibitem{zhu2022nice}
Zihan Zhu, Songyou Peng, Viktor Larsson, Weiwei Xu, Hujun Bao, Zhaopeng Cui,
  Martin~R Oswald, and Marc Pollefeys.
\newblock Nice-slam: Neural implicit scalable encoding for slam.
\newblock In {\em IEEE CVPR}, pages 12786--12796, 2022.

\end{thebibliography}
